\documentclass[sigconf]{acmart}

\AtBeginDocument{%
  }

\copyrightyear{2024}
\acmYear{2024}
\setcopyright{rightsretained}
\acmConference[CHI '24]{Proceedings of the CHI Conference on Human Factors in Computing Systems}{May 11--16, 2024}{Honolulu, HI, USA}
\acmBooktitle{Proceedings of the CHI Conference on Human Factors in Computing Systems (CHI '24), May 11--16, 2024, Honolulu, HI, USA}
\acmDOI{10.1145/3613904.3642767}
\acmISBN{979-8-4007-0330-0/24/05}

\usepackage{xspace}
\usepackage{enumitem}
\usepackage{subfig}
\usepackage{xcolor}
\usepackage{pifont}

\definecolor{ForestGreen}{RGB}{34,139,34}
\newcommand{\cmark}{\textcolor{ForestGreen}{\ding{51}}}%
\newcommand{\xmark}{\textcolor{red}{\ding{55}}}%
\newcommand{\tmark}{\textcolor{orange}{\ding{115}}}%

\definecolor{Issue1}{HTML}{D03416}
\definecolor{Issue2}{HTML}{13768E}
\definecolor{Issue3}{HTML}{9B4F0F} 
\definecolor{Issue4}{HTML}{008000}
\definecolor{Issue5}{HTML}{C99E10}

\newcommand{\system}{AccessLens\xspace}

\newif\ifsubmit
\submittrue

\ifsubmit
\newcommand{\topic}[1]{}
\newcommand{\nk}[1]{}
\newcommand{\jk}[1]{}
\newcommand{\el}[1]{}
\newcommand{\mhq}[1]{}

\newcommand{\IssueOneA}[1]{{#1}}
\newcommand{\IssueOneB}[1]{{#1}}
\newcommand{\IssueOneC}[1]{{#1}}
\newcommand{\IssueTwoAtoC}[1]{{#1}}
\newcommand{\IssueTwoA}[1]{{#1}}
\newcommand{\IssueTwoB}[1]{{#1}}
\newcommand{\IssueTwoC}[1]{{#1}}
\newcommand{\IssueThreeA}[1]{{#1}}
\newcommand{\IssueThreeB}[1]{{#1}}

\newcommand{\IssueFourA}[1]{{#1}}
\newcommand{\IssueFourB}[1]{{#1}}
\newcommand{\IssueFourC}[1]{{#1}}

\newcommand{\IssueFive}[1]{{#1}}

\else

  \newcommand{\topic}[1]
{{\fontfamily{cmss}\selectfont \bfseries \color{blue} Topic: #1}}
 \newcommand{\nk}[1]
{{\fontfamily{cmss}\selectfont \bfseries \color{red} NK: #1}}
 \newcommand{\jk}[1]
{{\fontfamily{cmss}\selectfont \bfseries \color{violet} JK: #1}}
\newcommand{\el}[1]
{{\fontfamily{cmss}\selectfont \bfseries \color{cyan} EL: #1}}

\newcommand{\mhq}[1]
{{\fontfamily{cmss}\selectfont \bfseries \color{magenta} MHQ: #1}}

\newcommand{\IssueOneA}[1]{\colorbox{Issue1}{\textcolor{white}{\#1A}} \textcolor{Issue1}{#1}}
\newcommand{\IssueOneB}[1]{\colorbox{Issue1}{\textcolor{white}{\#1B}} \textcolor{Issue1}{#1}}
\newcommand{\IssueOneC}[1]{\colorbox{Issue1}{\textcolor{white}{\#1C}} \textcolor{Issue1}{#1}}
\newcommand{\IssueTwoAtoC}[1]{\colorbox{Issue2}
{\textcolor{white}{\#2AtoC}} \textcolor{Issue2}{#1}}
\newcommand{\IssueTwoA}[1]{\colorbox{Issue2}
{\textcolor{white}{\#2A}} \textcolor{Issue2}{#1}}
\newcommand{\IssueTwoB}[1]{\colorbox{Issue2}{\textcolor{white}{\#2B}} \textcolor{Issue2}{#1}}
\newcommand{\IssueTwoC}[1]{\colorbox{Issue2}{\textcolor{white}{\#2C}} \textcolor{Issue2}{#1}}
\newcommand{\IssueThreeA}[1]{\colorbox{Issue3}{\textcolor{white}{\#3A}} \textcolor{Issue3}{#1}}
\newcommand{\IssueThreeB}[1]{\colorbox{Issue3}{\textcolor{white}{\#3B}} \textcolor{Issue3}{#1}}

\newcommand{\IssueFourA}[1]{\colorbox{Issue4}{\textcolor{white}{\#4A}} \textcolor{Issue4}{#1}}
\newcommand{\IssueFourB}[1]{\colorbox{Issue4}{\textcolor{white}{\#4B}} \textcolor{Issue4}{#1}}
\newcommand{\IssueFourC}[1]{\colorbox{Issue4}{\textcolor{white}{\#4C}} \textcolor{Issue4}{#1}}

\newcommand{\IssueFive}[1]{\colorbox{Issue5}{\textcolor{white}{\#5}} \textcolor{Issue5}{#1}}

\fi

\begin{document}


\title[\system]{AccessLens: Auto-detecting Inaccessibility of Everyday Objects}



\author{Nahyun Kwon}
\email{nahyunkwon@tamu.edu}
\affiliation{%
    \institution{Texas A\&M University}
    \city{College Station}
    \state{Texas}
    \country{USA}
}

\author{Qian Lu}
\email{qianlu@tamu.edu}
\affiliation{%
    \institution{Texas A\&M University}
    \city{College Station}
    \state{Texas}
    \country{USA}
}

\author{Muhammad Hasham Qazi}
\email{qazihasham@gmail.com}
\affiliation{%
    \institution{Texas A\&M University}
    \city{College Station}
    \state{Texas}
    \country{USA}
}

\author{Joanne Liu}
\email{joannechanliu@tamu.edu}
\affiliation{%
    \institution{Texas A\&M University}
    \city{College Station}
    \state{Texas}
    \country{USA}
}

\author{Changhoon Oh}
\email{changhoonoh@yonsei.ac.kr}
\affiliation{%
    \institution{Yonsei University}
    \city{Seoul}
    \state{}
    \country{Korea}
}

\author{Shu Kong}
\email{skong@um.edu.mo, shu@tamu.edu}
\affiliation{%
    \institution{University of Macau}
    \city{Texas A\&M University}
    \state{}
    \country{}
}


\author{Jeeeun Kim}
\email{jeeeun.kim@tamu.edu}
\affiliation{%
    \institution{Texas A\&M University}
    \city{College Station}
    \state{Texas}
    \country{USA}
}

\renewcommand{\shortauthors}{Kwon et al.}

\begin{abstract}

In our increasingly diverse society, everyday physical interfaces often present barriers, impacting individuals across various contexts. This oversight, from small cabinet knobs to identical wall switches that can pose different contextual challenges, highlights an imperative need for solutions. Leveraging low-cost 3D-printed augmentations such as knob magnifiers and tactile labels seems promising, yet the process of discovering \textit{unrecognized} barriers remains challenging because disability is context-dependent. We introduce AccessLens, an end-to-end system designed to identify inaccessible interfaces in daily objects, and recommend 3D-printable augmentations for accessibility enhancement. Our approach involves training a detector using the novel AccessDB dataset designed to automatically recognize 21 distinct Inaccessibility Classes (e.g., bar-small and round-rotate) within 6 common object categories (e.g., handle and knob). AccessMeta serves as a robust way to build a comprehensive dictionary linking these accessibility classes to open-source 3D augmentation designs. Experiments demonstrate our detector's performance in detecting inaccessible objects.

\vspace{-2mm}
\end{abstract}


\vspace{-4mm}

\begin{teaserfigure}
\centering
  \includegraphics[width=\textwidth]{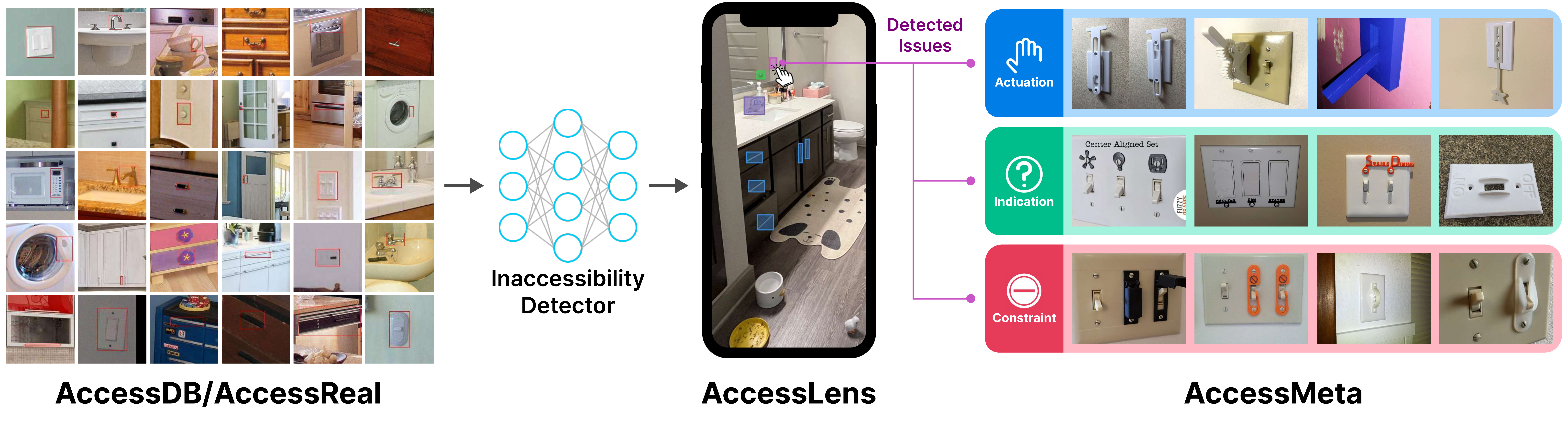}
  \vspace{-7mm}
  \caption{
  AccessLens system overview. AccessLens provides a mobile toolkit to scan indoor scenes and detect inaccessibility in everyday objects. Inaccessibility detection is developed on our dataset AccessDB and AccessReal, consisting of indoor scene images annotated with inaccessibility classes on daily objects. We contribute AccessMeta, a metadata that categorizes 3D assistive designs, enabling auto-suggestions to improve daily accessibility. 
 }
  \Description{The figure shows a flowchart from left to right. On the left, there is a 5 by 6 grid of images with a bounding box around an item in each image, all from AccessDB and AccessReal. The grid then points to a network icon labeled Inaccessibility Detector, which points to a mobile phone screen that has an image of a bathroom, with bounding boxes around several items in the photo. One of the bounding boxes around the outlet on the wall has an arrow pointing to another diagram that has assistive designs from AccessMeta sorted in three rows: Actuation, Indication, and Constraint from top to bottom. }
  \vspace{2mm}
  \label{fig:teaser}
\end{teaserfigure}
\maketitle

\section{Introduction}

While the traditional definition of disability has revolved around individuals' varied abilities, understanding disability as \textit{`mismatched interactions'}~\cite{microsoft_inclusive_design} emphasizes diverse contexts that can create barriers within environments.
Consider someone with a wrist injury struggling with everyday tasks like opening a water bottle or using a toothbrush single-handedly; new parents suddenly recognize potential hazards at home, such as electric outlets.
However, recognizing such contextual disability and proactively rectifying them remains challenging for inexperienced users because they use prior experiences that could be biased.
It is non-trivial to foresee unfamiliar interaction scenarios (e.g., managing everyday tasks by being one-handed), leading them to cope with difficulties without promptly addressing interaction challenges.

\IssueOneB{
\textit{``If the design is accessible, everyone benefits''}~\cite{UniversalDesign}; the accessibility community has highlighted the importance of engaging everyone in improving accessibility. 
Traditional approaches to raising awareness and fostering proactive efforts focused on cultivating empathy and mutual understanding among non-disabled individuals. The goal was to evoke recognition of unnoticed discomfort inherent in daily interfaces, particularly from the perspective of individuals with disabilities~\cite{teaching-a11y,teachingempathy, blindfold-empathy, myday-wheelchair, labs-learning-a11y}.
However, these approaches had inherent limitations in simulating disabilities, which could inadvertently lead to biases and cognitive gaps against individuals without disabilities~\cite{unintended-negative-simulations}. 
Although well-structured textual guidelines and compliances~\cite{ada_accessible, ibc, aarp_homefit_guide} encompass exhaustive domain knowledge from experts, those remain static, exclaiming the need for interactive systems. 
However, while the disability is context-dependent, implying that anyone can experience challenges without permanent disability, the latest AI-powered interactive tools~\cite{homefit_ar, rassar2022} predominantly focus on specific target groups, such as wheelchair users or older adults, missing the contextual variances, i.e., temporary and situational cases~\cite{microsoft_inclusive_design}. 
Moreover, many solutions entail renovation or replacements, which is often costly thus mentally burdening, limiting the practicality/applicability of existing tools in promoting pro-social behaviors. 
}
There remain three major user challenges:
\begin{itemize}[noitemsep, topsep=0pt]
    \item Which objects are inaccessible?
    \item Why and when do these objects become inaccessible?
    \item \IssueOneA{How can a user without prior experiences identify them and find appropriate solutions?}
\end{itemize}
 
\begin{figure}[!t]
    \centering
    \includegraphics[width=\linewidth]{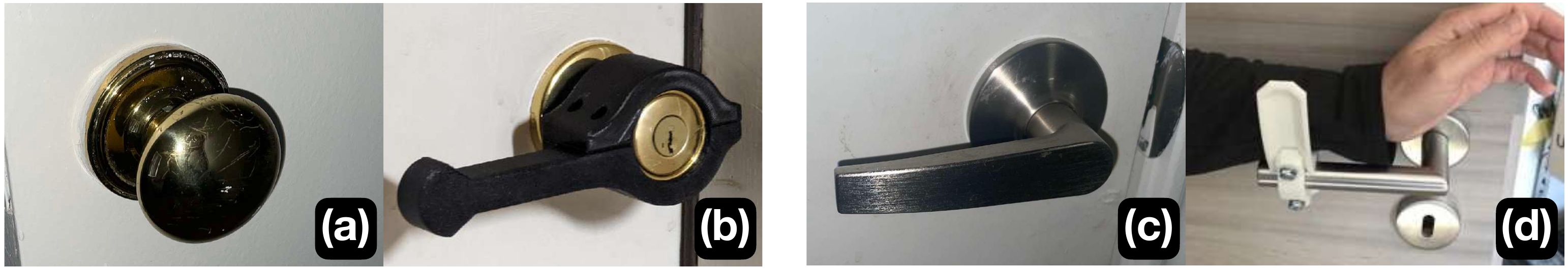}
    \vspace{-3mm}
    \caption{(a) A round knob's accessibility can be improved by (b) lever extension~\cite{door_lever_extension} while (c) a lever handle's accessibility is improved by an (d) arm extension ~\cite{door_arm_extension}. Everyday objects portray different accessibility barriers to people under different contexts. }
    \Description{The figure shows four images. Image (a) and image (b) are paired together and show an image of a round door knob and a lever extension that is wrapped around the knob and sticking out to the left side, resembling a door handle. Images c and d are paired, with the image (c) showing an image of (a) lever handle and image d showing an example of an arm extension. The extension is attached at the end of the handle and an arm is placed between the door and handle, where the wrist is rested on the curved piece of the extension.}
    \label{figure:doorknob}
\vspace{-2mm}
\end{figure}

We introduce \textbf{AccessLens}, an end-to-end system to automate detecting contextual barriers from everyday objects, and suggest 3D-printed assistive augmentations. Figure~\ref{fig:teaser} shows system overview. 
AccessLens is built upon novel datasets, \textbf{AccessDB/AccessReal} to train inaccessibility detectors, and \textbf{AccessMeta}, metadata to understand interaction types and required human capabilities of physical objects presented as their interaction attributes.
As existing datasets (e.g., \cite{COCO, PASCAL, ade20k}) with indoor scene images do not articulate inaccessibility to automate detection, AccessDB was built to imbue accessibility knowledge using 21 Inaccessibility Classes (IC). 
Designed to foster understanding of how 3D assistive augmentations can resolve contextual disabilities, AccessMeta provides the link between 3D augmentations and interaction types/contexts of existing objects, such as a lever extension for a door knob that removes sophisticated motor skills (Figure~\ref{figure:doorknob}a-b) and an arm-pull extension for a lever for an alternative operation (Figure~\ref{figure:doorknob}c-d). 

In sum, our contributions are three-fold:

\begin{itemize}
\item \IssueFourB{A holistic survey of large-scale 3D assistive augmentations in online repositories and understanding of their interaction properties, resulted in \textbf{AccessMeta}, a metadata to auto-classify them;}
\item \textbf{AccessDB \& AccessReal}: A dataset for auto-detection of inaccessible objects and parts from indoor scenes with 10k annotated objects under 21 Inaccessibility Classes with realistic high-res dataset for testing;

\item \textbf{AccessLens}: End-user system to detect inaccessibility and to obtain design recommendations through 3D printed augmentations to update legacy objects
\end{itemize}

\begin{figure}[!b]
    \centering
    \includegraphics[width=\linewidth]{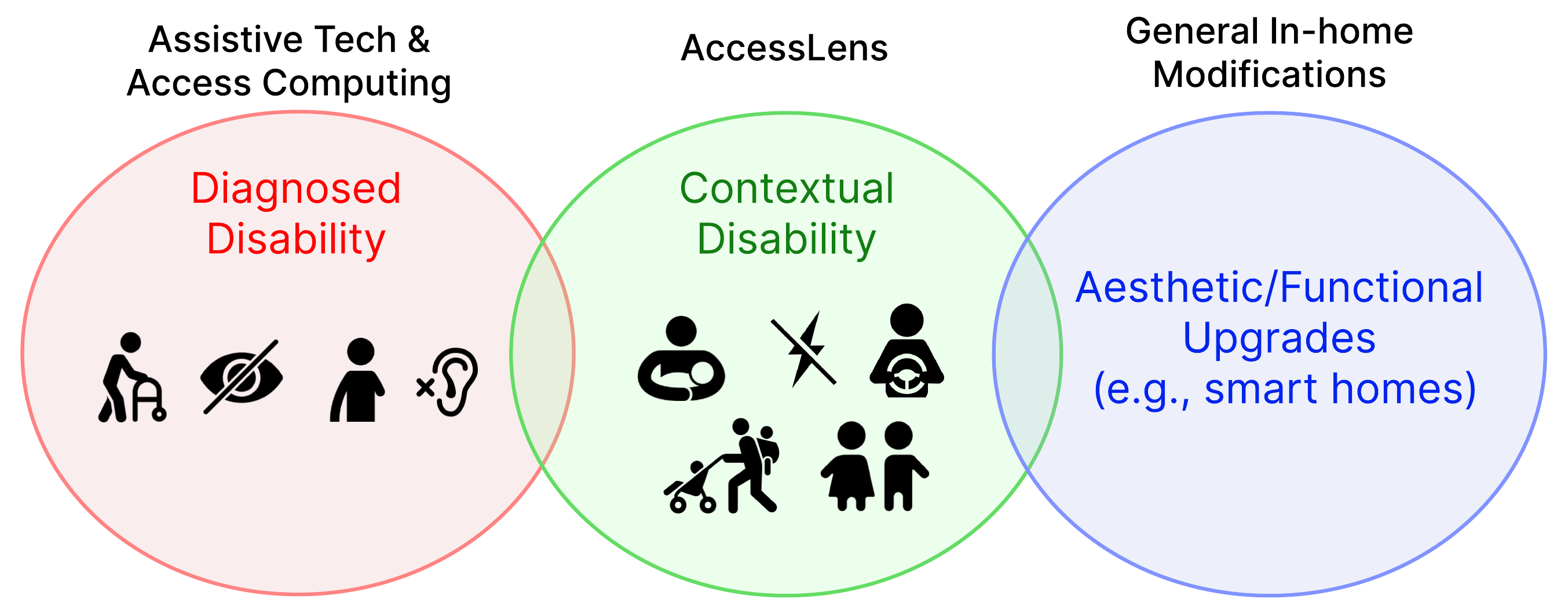}
    \vspace{-3mm}
    \caption{\IssueOneA{AccessLens's target user scope compared to existing assistive technique works and general in-home modifications. AccessLens supports users with limited awareness but who can easily become disabled under various contexts.}}
    \Description{This figure shows a 3-part Venn Diagram with the categories labeled from left to right `Assistive Tech and Access Computing', `AccessLens', and `General in-home Modification'. Under `Assistive Tech and Access Computing' is a circle that is labeled Diagnosed Disability with icons that represent common disabilities such as mobility, visual, and hearing impairment. The center circle, under `AccessLens' is labeled `Contextual Disability' with icons such as someone carrying a baby, pushing a stroller, and a shock hazard symbol. The last circle under `General in-home Modification' has written text that says `Aesthetic/Functional Upgrades (e.g. smart homes)'.}
    \label{figure:target-users}
\end{figure}

\IssueTwoAtoC{We evaluate our contributions through user studies and technical experiments. First, a preliminary user evaluation of the AccessLens system prototype helps understand how AccessLens enhances awareness and willingness to take pro-social behaviors. Second, we assess an end-to-end pipeline---capturing the indoor environment to retrofitting 3D augmentations---with inexperienced users and two experts in assistive technology.
The evaluation of AccessMeta engaged crowdworkers in annotating the dictionary with 280 3D augmentations. We also evaluate AccessDB/AccessReal with off-the-shelf detectors.}

Our vision for AccessLens is to empower \IssueOneA{users with limited awareness to recognize hidden daily accessibility challenges thus to be more attentive to daily challenges under diverse contexts and extents. AccessLens does not require diagnosed disability, prior experience, and domain expertise to recognize inaccessibility. Figure~\ref{figure:target-users} shows our scope on target demographics compared to existing approaches.}

\begin{table*}[h]
\begin{tabular}{|c|c|c|c|c|c|c|}
\hline
& \begin{tabular}[c]{@{}c@{}}Compliances/\\ guidances\\e.g., \cite{ada_accessible, aarp_homefit_guide}\end{tabular} & \begin{tabular}[c]{@{}c@{}}MS Inclusive\\ Guidebook\\\cite{microsoft_inclusive_design}\end{tabular} & \begin{tabular}[c]{@{}c@{}}Project\\Sidewalk\\\cite{project-sidewalk}\end{tabular} & \begin{tabular}[c]{@{}c@{}}Homefit AR\\\cite{homefit_ar}\end{tabular} & \begin{tabular}[c]{@{}c@{}}RASSAR\\\cite{rassar2022} \end{tabular}& \begin{tabular}[c]{@{}c@{}}AccessLens\\ (ours)\end{tabular} \\ \hline
Interactive                                                                & \xmark                                                            & \xmark                                                            & \cmark                                                        & \cmark      & \cmark  & \cmark                                                       \\ \hline
Indoor accessibility                                                                & \cmark                                                            & \tmark                                                            & \xmark                                                        & \cmark      & \cmark  & \cmark                                                       \\ \hline
Contextual disability                                                               & \xmark                                                            & \cmark                                                            & \xmark                                                        & \xmark      & \cmark  & \cmark                                                       \\ \hline
\begin{tabular}[c]{@{}c@{}}Auto detection\end{tabular}             & \xmark                                                            & \xmark                                                            & \cmark                                                        & \cmark      & \cmark  & \cmark                                                       \\ \hline
\begin{tabular}[c]{@{}c@{}}Interaction type detection\end{tabular} & \xmark                                                            & \xmark                                                            & \xmark                                                        & \tmark      & \xmark  & \cmark                                                       \\ \hline
Low-cost adapatations                                                               & \xmark                                                            & \xmark                                                            & \xmark                                                        & \xmark      & \tmark  & \cmark                                                       \\ \hline
\end{tabular}
\caption{\IssueOneC{Position of AccessLens compared against prior works.}}
\Description{The table compares the position of AccessLens with other works. The different aspects it is compared in are the rows while the columns are the different prior works including AccessLens. The table has check marks and cross symbols to represent whether the column satisfies each row.}
\label{tab:positioning}
\end{table*}

\section{Related Work}

\subsection{Interactive Tools to Evaluate Accessibility}

\IssueOneB{There exist numerous standards and normative tools to help non-experts learn cumulative knowledge.} The Americans with Disabilities Act (ADA) Standards for Accessible Designs~\cite{ada_accessible} and the International Building Code (IBC)~\cite{ibc} represent comprehensive frameworks to alleviate mobility challenges.
Increasing interests are in their interactivity,
for instance, improving indoor access for older adults through interactive systems~\cite{homefit_ar, buzzelli2020vision, luo2018vision, easom2020inhouse, luo2017computer}.  
Homefit AR~\cite{homefit_ar, aarp_homefit_guide} guides users through questionnaires to precisely locate issues with object types and recommends alternatives with better access.
The closest prior work of ours is RASSAR~\cite{rassar2022}, a mobile AR application to assess indoor accessibility upon standards such as low tables, narrow entryways, and dangerous items exposed.
\IssueOneA{While these works respond to the needs of special interest groups (e.g., older adults and wheelchair users), a broader population is often excluded, since they have not experienced disabilities and could overlook contextual or situational disabilities.}
We are to provoke solutions with an emphasis on the engagement of a more diverse community in creating accessible and accommodating indoor spaces.

\subsection{Advancing Accessibility: Beyond Empathy and Simulations} 
Fostering empathy is discussed in many disability studies to elevate awareness about the lived experiences of disabled people~\cite{teaching-a11y, teachingempathy, PromiseofEmpathy}.
While simulating disabilities such as blindfolding~\cite{blindfold-empathy}, having colorblind effects~\cite{labs-learning-a11y}, or trying wheelchairs~\cite{myday-wheelchair} has gained popularity, disability advocates disparage simulated disability~\cite{PromiseofEmpathy, empathy-cannot-sustain}; as it is difficult to accurately replicate the real experiences~\cite{infusing-disability-simulations}.
Empathy alone may not suffice to sustain attention~\cite{empathy-cannot-sustain}, simulations may inadvertently create biases or distress~\cite{unintended-negative-simulations}, resulting in perpetuated ableism~\cite{hofmann2020living}.
More recent focus is on co-designing \textit{with} people with disabilities (e.g., \cite{makehealth, pathfinder, bentomuseum, interdependence}); e.g., citizens, healthcare professionals, and designers co-design personalized healthcare solutions ~\cite{makehealth}, sighted and blind participants design building navigation together~\cite{pathfinder}.
Collective efforts to enhance the user experience can extend the impact beyond individuals with disabilities alone, encompassing different abilities of all ~\cite{why-accessible-for-everyone, microsoft_inclusive_design, we-all-benefit}. 
Unfortunately, there have been only little to no systems to engage non-experienced users to cultivate inclusivity.
Our approach is motivated by ``\textit{design for one, expand to all}'' and ``\textit{learning from adaptations}''~\cite{microsoft_inclusive_design} to promote awareness through noticing and better designs.

\subsection{Indoor Scene Understanding}
Visual perception of indoor places is a critical first step to improving one's quality of life \cite{savage2022robots}.
Various datasets were released to train detectors.
Some early datasets such as MIT indoor scenes~\cite{quattoni2009recognizing} and SUN RGB-D~\cite{song2015sun} have advanced techniques to train recognition models.
Synthetic datasets such as HyperSim~\cite{roberts2021hypersim} can further advance recognition models with their variety and quantity.
While there exist relevant datasets such as Gibson~\cite{xia2018gibson}, offering a virtual visual navigation platform, PartNet~\cite{mo2019partnet} with focus on part of indoor objects, and
BEHAVIOR-1K~\cite{BEHAVIOR-1K}, data for embodied AI systems to foster human-robot interaction, none have centered interaction types to assess their accessibility and user contexts.
We find ADE20K~\cite{ade20k}, which is a large-scale indoor scene dataset with hierarchical annotations of objects in images at the pixel level, promising.
Refining the hierarchical taxonomy of objects and parts by ADE20K includes object categories and parts, we curate datasets by re-annotating potentially inaccessible objects to train and evaluate inaccessibility detectors.

\subsection{3D-Printed Augmentations to Improving Access to Legacy Objects}

While it is not feasible to replace \textit{all} existing objects overnight \cite{arabi2022mobiot, li2019robiot}, 3D-printed assistive designs~\cite{buehler2015sharing, ashbrook2016towards} promises low-cost, custom solutions to redress everyday interaction challenges (e.g., \cite{chen2016reprise, Guo:2017:Facade, buehler2015sharing}).
These adaptations can range from magnifying cabinet knobs for improved grip (e.g., `ThisAbles'~\cite{university2018thisAbles}) to self-serving medicine dispensers~\cite{arabi2022mobiot}.
Similar to the modular approach employed in the modern software engineering paradigm, wherein updates are selectively applied only where changes are necessary~\cite{niu2014rockjit}, the augmentation allows for unit-by-unit enhancements tailored to specific needs. Barriers to 3D printing have been significantly lowered~\cite{berman2020anyone}, existing works studied motivations behind online communities sharing assistive 3D designs ~\cite{buehler2015sharing} and proposed computational customization solutions (e.g., ~\cite{chen2016reprise}). 
While documents based on similarity can classify shared designs' objectives~\cite{customizAR}, current search relies on designer-created descriptions, \IssueOneA{often failing users to explore viable designs to rectify hidden inaccessibility that is not obvious to those without diagnosed disabilities.}
Discovering suitable designs heavily relies on keyword-based searches, relying on the textual information provided by the authors: titles, descriptions, and tags only.
We propose novel metadata to categorize existing 3D assistive augmentations for better identification of solutions.

\IssueOneC{In sum, Table~\ref{tab:positioning} summarizes the position of Accesslens. }




\begin{figure*}[h]
    \centering
\includegraphics[width=1\linewidth]{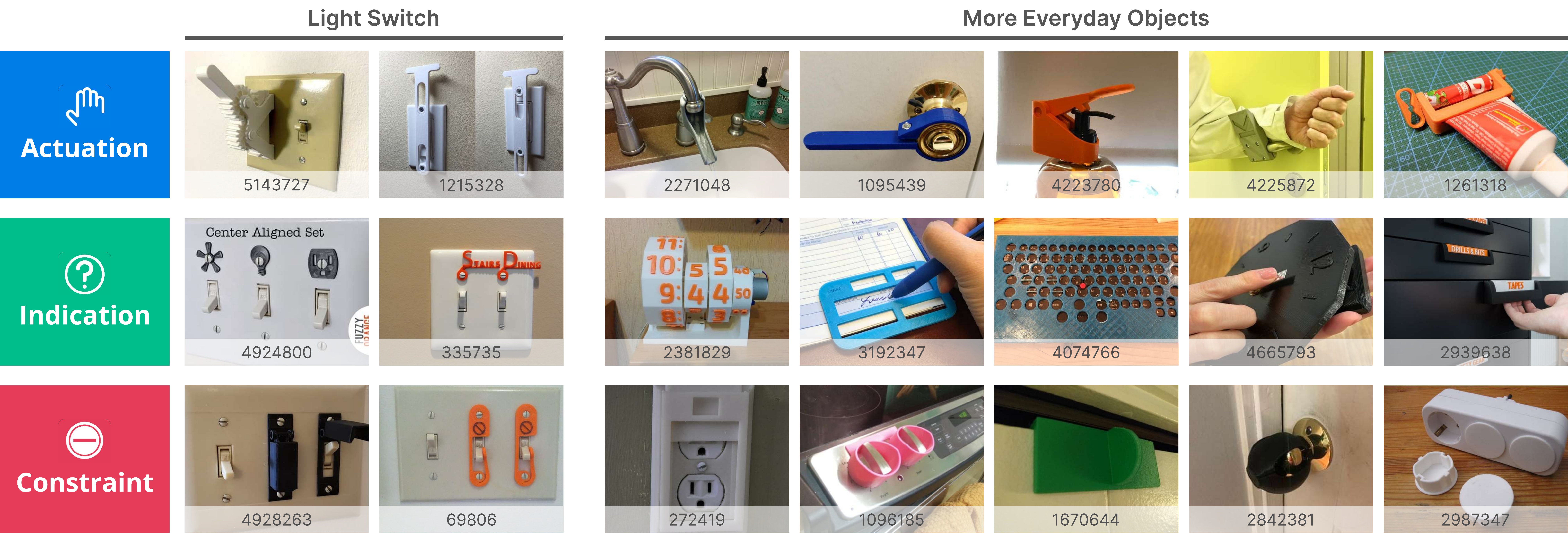}
\vspace{-5mm}
    \caption{Examples of 3D assistive augmentations that belong to three categories, obtained from our in-the-wild survey with iterative affinity diagramming. Each design has a \textit{thing\_id} at the bottom, and the design page can be located at https://www.thingiverse.com/thing:\textit{thing\_id}. Examples show that various challenges, such as motor and sensory barriers, can present even for one object. 3D augmentations are actively used to address challenges without requiring total replacement. 
    }
    \Description{The figure shows a 3-row grid of 21 images of assistive designs categorized into actuation, indication, and constraint from up to down. The left two columns are assistive designs used on light switches. The first column, from top to bottom, shows a key-shaped attachment angled upwards that makes the light switch longer and easier to push up and down, a light-switch panel with stickers to indicate the purpose of each switch in the second row, along with a plastic cover that blocks users from accessing the light switch in the third row. The right columns are other examples of assistive designs used in more everyday objects, including a toothpaste squeezer, labels to put on drawer handles, and a table corner cover.}
    \label{fig:augmentations}
\end{figure*}

\section{Designing AccessLens}

We introduce two design studies;
\IssueThreeA{the first investigates how people currently make changes to their environments by adopting 3D printed augmentations through an in-the-wild survey. 
Taking account into the design objectives and interactions that entail, the second examines our design probe with 8 participants whether the system helps naive users interpret accessibility challenges differently.
}

\subsection{Design Study \#1. Understanding Interaction Contexts: In-the-Wild Survey}
\label{sec:design-2}

We conducted an exploratory survey on Thingiverse~\cite{thingiverse}, to gain insights into why individuals are motivated to create 3D assistive augmentations and modify existing physical objects to address specific contextual or situational interaction challenges. 
First, we listed several indoor objects that are very common around us, including door knobs, light switches, etc.
Then we retrieved 3D designs that are for those objects, indicating 3D designs tend to augment targeted real-world objects from Thingiverse. 
We employed an iterative process of affinity diagramming, which was collaboratively performed by four of our authors. In the affinity diagramming process, we classified augmentations considering three primary criteria: (1) their intended objective, which refers to the barriers the augmentations aim to address, (2) the type of objects the augmentations target, and (3) any related motions or actions associated with their use.
Our empirical findings revealed that even for objects that are under the same class (e.g., door knob/handle, light switch), the augmentations are much more diverse due to differences in the object's type (e.g., single toggle light switch vs. rocker switch). 
This diversity emanates from shapes, motions, and objectives, which inspired us to develop AccessDB, our refined dataset with inaccessibility classes of AccessMeta. 
This iterative affinity study resulted in three high-level functions of adaptations as follows and example augmentations are shown in Figure~\ref{fig:augmentations}.

\begin{itemize}[noitemsep, leftmargin=*]
    \item \textbf{Reducing motor requirements, change needed motion types [Actuation]:} Designs that shift types of motions needed to operate (e.g., rotation to linear push) or reduce workload (e.g., reduce required power to manipulate interfaces, or allow one hand instead of two hands); for people with motor limitations.
    \item \textbf{Furnishing with visual/tactile cues [Indication]:} Designs that create multi-modal functions for identification, providing labels (e.g., switch identifiers, toggling sound); for people with sensory limitations.
    \item \textbf{Adding constraints [Constraint]:} Designs that prevent a targeted population from operating a task by limiting their operation mainly due to safety reasons (e.g., cabinet lock, switch lock, stove knob stopper); for people with cognitive limitations or child-access/child-proof products.
\end{itemize}



\subsection{Design Study \#2. Design Probe} \label{sec:preliminary-evaluation}
\begin{figure*}[ht]
    \centering
    \vspace{-2mm}
    \includegraphics[width=0.95\textwidth]{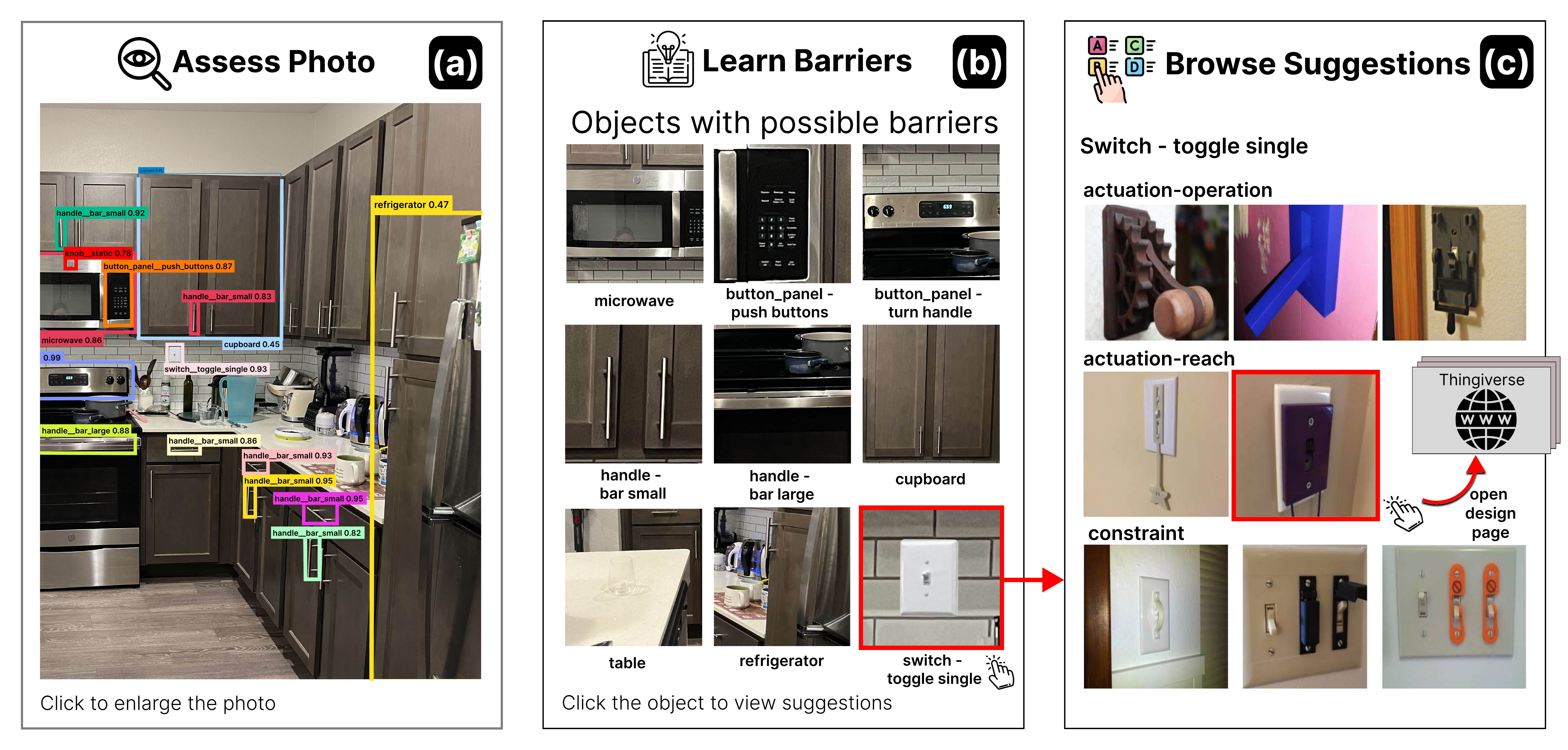}
    \vspace{-5mm}
    \caption{AccessLens prototype overview. AccessLens allows users to scan an uploaded photo (a), view the detected inaccessible objects (b), and upon a click of a detected object, browse through the available suggestions (c).}
    \Description{The figure displays three views from the AccessLens website, arranged from left to right: (a) Assess Photo, (b) Learn Barriers, and (c) Browse Suggestions. On the left, the `Assess Photo' view presents a kitchen photo with bounding boxes marking various objects and a note suggesting users click for a larger view. The middle `Learn Barriers' view has a 3-by-3 grid of zoomed-in objects with possible barriers detected from the kitchen. A light switch image is highlighted with a cursor icon indicating selection. The right-most `view that shows three rows of assistive design suggestions for the light switch, labeled "actuation-operation," "actuation-reach," and "constraint." In the "actuation-reach" row, there is a yellow box around one of the images, The cursor icon to represent when clicking on the image will lead to the open web page of the clicked design in "Thingiverse". }
    \vspace{-2mm}
    \label{figure:demo}
\end{figure*}

We developed the prototype of the AccessLens and conducted a comparative study to assess its validity and advanced features over the baseline, MS Inclusive Design Guidebook~\cite{microsoft_inclusive_design}. \IssueThreeB{Compared to other normative tools that are targeted to diagnosed disabilities, e.g., ADA Standards for Accessible Design~\cite{ada_accessible}, the MS guidebook is the foremost design guideline that argues accessibility as a universal daily challenge \textit{for all}, encouraging recognizing exclusion, extending the inaccessibility concept to contextual from a permanent problem. Herein, the disability is discussed not as a personal health condition, but as \textit{`mismatched human interaction'} which we see the potential to rectify through augmentations. 
Thinking of solutions for those situational disabilities can allude to a design for one that can benefit all~\cite{why-accessible-for-everyone}, empowering people to learn from diversity}.
AccessLens prototype (Figure~\ref{figure:demo}) includes objects with detected potential accessibility challenges. Tapping on objects, the system displays relevant 3D augmentations depending on contextual needs. 
We provide the contexts through a catalog approach, helping users learn from viewing adaptations list, which also presents design implications to nurture people's understanding of solutions. 

\vspace{-1mm}
\subsubsection{Participants.}
We recruited 8 participants from various backgrounds, including researchers who are not in the accessibility domain (\textit{N}=5), educators (middle/high school teacher, college professor, \textit{N}=3). Two self-identified as older adults (\textit{N}=2).
\IssueOneA{Aligning with our target users who do have limited experiences in the accessibility concepts, we recruited participants without diagnosed disabilities nor knowledge of accessibility study. We observed whether AccessLens promotes \textit{``thinking about daily inaccessibility''.}}

\subsubsection{Procedure.}
We chose a within-subject study.
We counter-balanced the conditions to reduce learning effects; half of the participants started with the baseline condition, and the other half started with the experimental condition.
The study sessions began with a pre-task interview. Participants then completed the same tasks under two conditions and finally, took a closing interview.
In the pre-task interview, participants shared their prior experiences when they encountered difficulties in interacting with everyday objects or witnessed someone else having issues. 
They were also asked if they had implemented any solutions to address such barriers.
One study condition is the Baseline condition, where participants access the link to the introduction video for MS Inclusive Design~\cite{ms-inclusive-video} and the MS Inclusive 101 guidebook~\cite{microsoft_inclusive_design} (MS guidebook, hereinafter). 
Participants were allowed to spend enough time reading the guidebook, without any time restrictions. 
During the task, participants were presented with indoor scene images and identified the objects that could present potential accessibility barriers. They were then asked to propose solutions. Subsequently, participants were asked to rate each suggestion on a 5-point Likert scale.
Participants were encouraged to use any necessary online resources (e.g., YouTube and Google Search) in the baseline.
In the experimental condition, participants used AccessLens but were not permitted to access other online resources. 
As shown in Figure~\ref{figure:demo}, the AccessLens displays indoor scene images of chosen, highlighted objects that could be inaccessible and offers applicable solutions. 
The task was repeated with a different indoor scene image.
After both conditions, a brief interview followed for 15 minutes.
We investigated their perceived usefulness by providing a survey questionnaire measuring three sub-metrics on a 5-Likert scale: (1) recognizing inaccessible objects, (2) comprehending related contexts, and (3) identifying solutions.
All responses and comments were documented for analysis.
The entire session took 1.5 hours on average, not exceeding 2 hours. The study has been approved by the institution's review board (IRB No.: IRB2023-0648)

\begin{figure}[h]
    \centering
    \includegraphics[width=\linewidth]{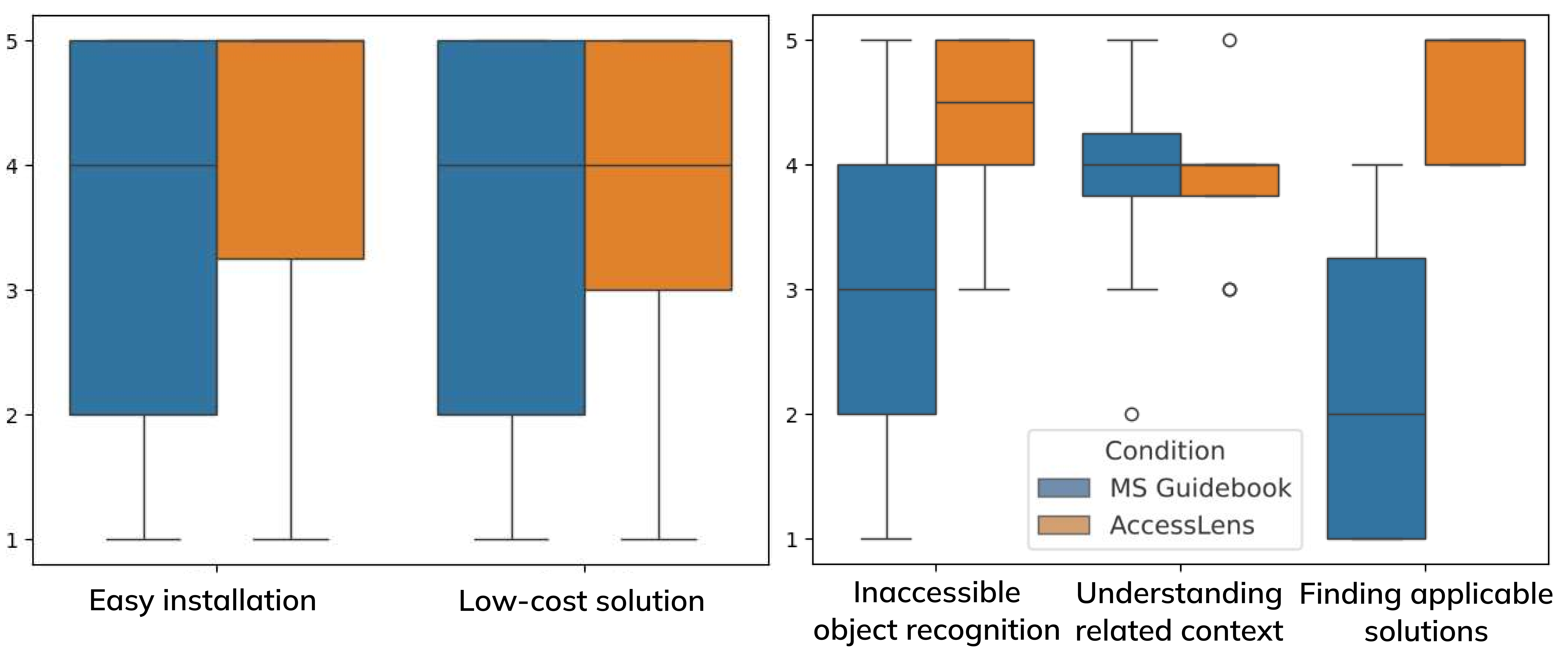}
    \vspace{-5mm}
    \caption{(a) 3D augmentation recommendations are assessed by two sub-metrics of easy installation and low-cost solution. (b) Perceived helpfulness is assessed by inaccessible object recognition, understanding contexts, and solution retrieval.}
    \Description{The figure shows two box plots side by side. For both, the y-axis value ranges from 0 to 5. (a) X-axis labels are easy installation and low-cost solutions for (a) and inaccessible object recognition, understanding related contexts, and finding applicable solutions for (b).}
    \label{figure:quant-metrics}
\end{figure}



\subsubsection{Findings \& Implications}
\hfill\\
\textbf{\#1. Ableism: Overlooked Inaccessibility and Gaps between noticing and an Action.}
P2 shared the story of their mother suffering from an ankle injury, leading her to stay seated at home until recovered. 
All often relied on family members for assistance, such as getting dressed with the help of a sibling (P2), and tried to circumvent challenges by struggling to use a non-dominant hand (P5) which was not perceived as a `disability' at that moment. 
\IssueOneA{Internalized ableism might explain this, where individuals may think disability \textit{``has to cross some threshold of difficulty or suffering to count''}~\cite{hofmann2020living} and do not think of their constraints as living disabilities to be addressed with solutions. Standing out to those who do not present diagnosed disabilities, ableism eventually misses the opportunities to renovate their environment for future contexts.}

\textbf{\#2. Learning from Adaptations.}
Participants noted that design recommendations make them infer contexts albeit no explicit descriptions were provided.
Several participants liked the persona spectrum presented in the MS guidebook, how different disabilities can relate to each other, broadening their understanding of disability. 
P5 mentioned that he now recalls he was temporarily disabled.
AccessLens achieved the same effect by cataloging various augmentations.
It encouraged participants to ``learn from adaptations''~\cite{microsoft_inclusive_design}.
Many were surprised by the variety of AccessLens recommendations, admitting they had not considered accessibility issues those designs could negate. \textit{``I hadn't thought these [could be an] issue before I saw these designs''} (P2, P4).
P3 felt gratitude for the detection \& suggestion together. \textit{``When I only saw the photo of the room [...], even when I see the detected objects, I didn't know which contexts it can pose barriers. When I saw the suggestions, I could imagine in which situations it can be helpful and what the objective is [of those or similar designs]''}.
We confirmed that presenting \textit{better designs} can inspire and let users comprehend the diversity.
P5 preferred AccessLens highlighting its transformative impacts; \textit{``We usually think only of the disabled [when we were asked to think about disability]. AccessLens makes me think that even the non-disabled can get help and apply the solutions in their environments''}.

\textbf{\#3. Mental load in \IssueOneA{Disability Accommodation.}}
We questioned the estimated installation expenditure under two sub-metrics: \textit{(1) easy installation} and \textit{(2) low-cost solution}.
Figure~\ref{figure:quant-metrics}a shows their estimation in easiness/affordability.
The average score does not indicate notable differences, possibly due to the learning effect; participants who experienced the AccessLens first tended to use their knowledge obtained during the following baseline condition.
Participants who began with the baseline guessed replacement or extensive renovations as the sole solutions. While perceived difficulty and cost varied among participants, they projected high cost and effort for replacements.
Most participants were curious about market products in the baseline condition. P5 imagined aggregated dials with small labels on a kitchen stove could be confusing for older adults, thinking individual knobs for each burner would be helpful, but questioning whether he could get one off-the-shelf.
In contrast, all found AcceeLens recommended 3D augmentations straightforward and cost-effective.
\textit{``I thought that we always needed complete replacements or renovations [...] Reviewing the suggestions, I realized that these solutions can be easily installed so I really want to install them, [e.g., childproof augmentations] to ensure safety''} (P3).

On the other hand, in baseline, none utilized external sources \IssueOneA{not knowing what and how to search, implying low engagement}. Only P1 tried general search keywords (e.g., assistive bathroom, accessible bathroom).
\textit{``I had to brainstorm to find the solutions [on my own]. Even with online resources allowed, I believe it wouldn't be that helpful because I need to know what to search for''} (P1).
This signifies the higher mental load keeps users from engaging in solution-seeking/adaptations.



\textbf{\#4. Written Guidebook vs. Interactive System.}
In the closing interview, participants evaluated two conditions across three sub-metrics: (1) the ability to recognize inaccessible objects, (2) understanding related contexts with barriers, and (3) retrieving applicable solutions. Figure~\ref{figure:quant-metrics}b summarizes participants' assessment, showing AccessLens outperforms the guidebook in terms of detecting inaccessible objects and seeking solutions. 

\begin{figure*}[h]
    \centering
    \includegraphics[width=0.98\textwidth]{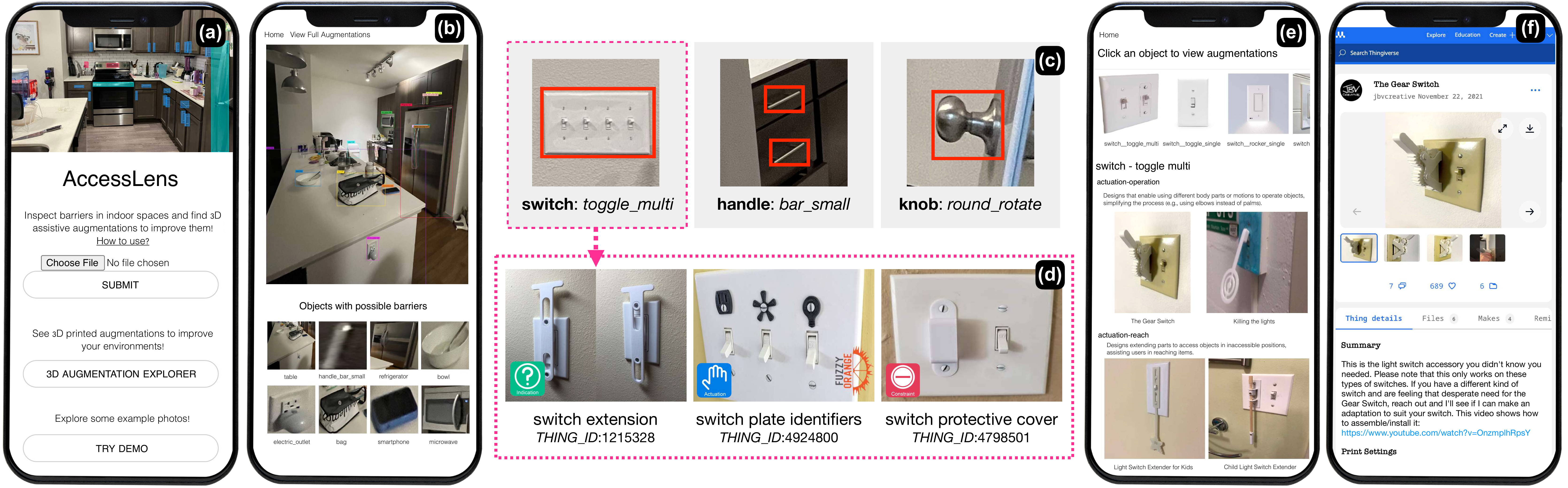}
    \caption{AccessLens: (a) main page, (b) an indoor image with detected objects with barriers, (c) example inaccessibility classes, (d) 3D-printed augmentations classified with AccessMeta, (e) a 3D augmentation explorer to view full suggestions, and (f) redirecting to the design page for details.}
    \Description{This figure illustrates the AccessLens user interface and system overview from left to right. The leftmost view is the photo upload view (a). (b) displays bounding boxes and the detected inaccessible objects in a captured photo. Users can select detected objects to navigate to the suggestion screen of the chosen object (b). The suggestion screen is right next to the summary view, it allows swiping through other detected objects horizontally, the figure shows three object examples: a toggle_multi style light switch, a small bar style handle, and a round rotate door knob (c). The system provides 3D printable assistive design suggestions in three categories: actuation, indication, and constraint, each accessed by clicking corresponding buttons. Three assistive design examples are given: a light switch extension for actuation, a light switch plate identifier for indication, and a light switch protective cover for constraint (d). (e) shows a design explorer page where users can view all possible designs. These designs are accessible on the Thingiverse website, tapping design directs users to the design webpage (f).}
    \label{figure:accesslens_ui_new}
\end{figure*}

\textbf{\#5. Interaction Design.}
In addition, participants desired, (1) implementing on a mobile reduces the user experience gap between capturing photos and inspection.
(2) Context-based filtering to reconcile accessibility evaluations to certain scenarios, increasing the system's versatility. 
(3) A summary view of all detection with bounding boxes to simplify the inspection process for a quick overview at a glance, and also to grasp the objectives of proposed accessibility enhancements swiftly. 
(4) Supplementary explanations for the categorization, i.e., AccessMeta categories, will enhance in-depth appreciation of the suggestions and their design intention. 
(5) A tutorial or instructional guide on how to capture photos would help users in providing clear and relevant images. 
These collective enhancements were reflected in AccessLens improvements.
We elaborate on an improved design as in Figure~\ref{figure:accesslens_ui_new}.
\vspace{-2mm}

\subsection{Design \& Implementation Considerations}
\label{design_considerations}

\textbf{Consideration \#1: One-shot Image Input} \label{sec:design-consideration-image-input}
From the HCI perspective, allowing users to upload a single photo of an indoor scene would offer a more pleasant experience, considering that our target users might not know where to focus.
While detection performance can benefit from multiple photos of the indoor scene, it is more friendly for users to take a single photo of the entire room or scanner view to check whether there exist any inaccessibility concerns. 
We target one-shot imagery of indoor scenes of interest as input, i.e., a panoramic scan of a bathroom, living room, and office space. 


\textbf{Consideration \#2. Semantic Understanding of Parts}
To assist users with different needs, \IssueFive{detecting part (doorknob from a door) and discerning the \textit{type} of the object (doorknob vs. lever) is critical to articulate contextual barriers beyond simple object detection.
The system must detect target objects and the parts where actual user interaction occurs, since each presents unique barriers with associated interaction types, for example, a knob for grab-pull vs. a knob for grab-rotate.} 
Therefore, the image dataset must contain indoor scenes with part-level annotations.

\textbf{Consideration \#3: Recognizing Disability Attributes.}
Various contexts change the way that people with a wide spectrum of capabilities interact with everyday objects; for a graphic designer wearing a splint due to chronic wrist pain, a door knob is not accessible as it requires hard grasping to rotate. 
People are often frustrated with a panel with identical toggle switches; without labels, they are forced to recall targets or try to get the right one turned on, sometimes causing safety breaches. 
 The disability context attributes of the objects
 might fortify the existing dataset.
In sum, 
\begin{enumerate}
    \item A user should be able to use a general view of scenes as input instead of a focused view of interested objects.
    \item The system must be able to semantically understand the detected objects (e.g., cabinet knob vs. door knob). 
    \item A new dataset must account for understanding various inaccessibility contexts beyond object/instance detection.
\end{enumerate}

\section{AccessLens}

AccessLens comprises AccessDB, AccessMeta, and the end-user toolkit, designed to seamlessly work together to assist end users in addressing accessibility challenges. The center of its functionality is AccessDB, a dataset used to train the inaccessibility detector, which analyzes images captured by users via a mobile user interface~\ref{fig:teaser}. The detector identifies inaccessible objects within various possible contexts. Leveraging AccessMeta, AccessLens suggests the design intentions and categories of 3D assistive augmentations. 

\subsection{AccessMeta: A Metadata of Assistive 3D-Printed Augmentations}
\label{sec:category}

We define \textit{``assistive augmentations''} herein as attachments to legacy physical environments, addressing inexplicit barriers in varying contexts. 
Ever since numerous 3D printing practitioners have open-sourced their creations online (c.f., \cite{buehler2015sharing}),
many were posted with voluntary textual descriptions with ``assistive'' to indicate the design intention.
Some not originally intended to be assistive missing relative tags could also be used for access but makes shopping through millions of designs by searching exhausting.
Navigating options is even more laborious due to ambiguity in language~\cite{customizAR}.
The structured rules or metadata to categorize assistive augmentations will broaden access to those designs, enabling users to explore easily.

\subsubsection{Coding corpus of assistive augmentations}
To tackle this, we surveyed large-scale data about designs on Thingiverse~\cite{thingiverse}, defining rules by observation such as retrieving relevant designs for target objects of interest.
As our goal is to assist users in searching 3D augmentations based on target objects in mind as approached similarly in prior works~\cite{chen2016reprise} and practice (e.g., ThisAble project \cite{university2018thisAbles}), we initiated our search with target objects, e.g., ``assistive door lever''.
While the existing categorization and corpus~\cite{buehler2015sharing} could be useful, designs classified under them do not necessarily represent augmentations. This also applies to CustomizAR taxonomy \cite{customizAR}, which primarily focuses on adaptive designs but assistive designs are only a small set. Consequently, we opted not to directly adopt this taxonomy in our corpus formation process.


We selected the initial search keywords of common indoor objects: \textit{door, drawer, cupboard, closet, outlet, light switch, switch, kitchen, utensil, cutlery, knife, spoon, fork, bottle, jar, bag, key, soap, shampoo, dispenser, nail clipper, can, pen, book, spray, phone, laptop, camera, toothbrush, toothpaste, clock, etc.}
We started observing the first 50 entries for their affinity defining the corpus.
Then we expanded the search, resulting in $\sim$1,600 entries by two sets of keywords overlap.
The first and second authors manually annotated their affinity by the common interaction types (Section~\ref{sec:design-2}), and the last author validated the results for agreement. 
With iterations and polishing, we define \textbf{AccessMeta}, the three high-level categories and their assistive functions, and common keywords and tags (Table~\ref{tab:accessmeta}). 

\begin{table}[h]

\begin{tabular}{c|c|l}
\toprule
\textbf{Category} & \textbf{Functions}                                        & \multicolumn{1}{c}{\textbf{\begin{tabular}[c]{@{}c@{}}Common keywords for \\ 3D assistive augmentations\end{tabular}}}      \\ 
\hline
\textbf{actuation}         & \begin{tabular}[c]{@{}c@{}}operation\\ reach\end{tabular}     & \begin{tabular}[c]{@{}l@{}}lever/hand extension, grip, \\ mount, opener holder/gripper, \\ string extension\end{tabular} \\ \hline
\textbf{constraint}        & limit access & \begin{tabular}[c]{@{}l@{}}cover, guard, protector, lock\end{tabular}                                                    \\ 
\hline
\textbf{indication}       & \begin{tabular}[c]{@{}c@{}}visual\\ tactile\end{tabular}      & label, identifier, tag        \\                                                \bottomrule
\end{tabular}
\caption{AccessMeta corpus to categorize 3D assistive augmentations. We found that the majority fall into three categories depending on their desired functions by augmenting real-world objects, often described by common keywords.
}
\vspace{-5mm}
\Description{The table organizes augmentation design categories into three rows, detailing their functions and associated keywords. The first row, labeled `actuation', is divided into `operation' and `reach' functions. `Operation' includes common keywords `level/hand extension', `string extension', and `opener'. `Reach' has common keywords `holder/gripper', `grip', and `mount'. The second row, labeled `constraint', has the function `limit access'. `Limit access' has common keywords `cover', `guard', `protector', and `lock'. The third row, labeled `indication', is divided into `visual' and `tactile' functions. `Visual' and `tactile' both have common keywords `label', `identifier', and `tag'.}
\label{tab:accessmeta}
\end{table}

\begin{figure*}[h]
    \centering
\includegraphics[width=1\textwidth]{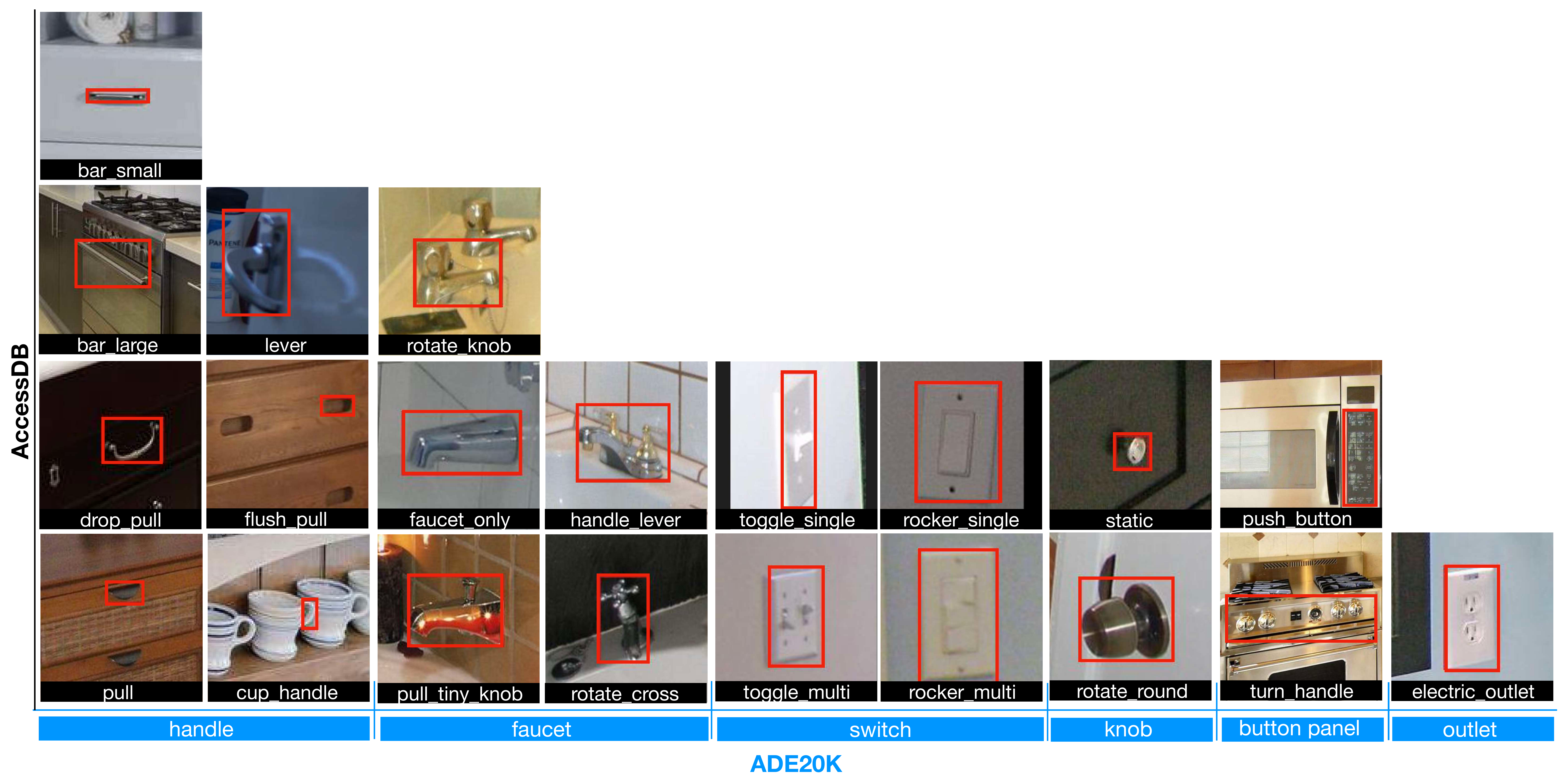}
\vspace{-6mm}
    \caption{We derive our AccessDB dataset by annotating indoor images from ADE20K dataset~\cite{ade20k} with 21 inaccessibility classes.
    We focus on 6 types of objects (blue-labeled names) which frequently appear to be inaccessible in daily life.
    }
\Description{The figure shows examples of 6 types of objects, which are handle, faucet, switch, knob, button panel, and outlet. The objects are from the ADE20K dataset and sorted into the inaccessibility classes that ultimately form our AccessDB dataset. Each image is a zoomed-in photo of the region around the inaccessible object and also has a red bounding box around the object. }
\label{figure:accessdb_labels}
\end{figure*}

{\textbf{(1) Actuation: Reduce motor requirements}}
refers to designs assisting people with operational difficulties (e.g., fine motor impairments, occupied hands) by extending or magnifying parts; including designs that reduce the required strength or alter the needed motion types.
Two functions are afforded if augmented: (help) \textit{operation} and \textit{reach}.
\textit{Actuation-operation} designs enable alternative operations using other body parts (e.g., elbow-push instead of hands-grab \& rotate) or motions or reduce needed power.
As an example, a doorknob extension (as in Figure~\ref{figure:doorknob}b) replaces the grasping-to-rotate with pushing-down.
Figure~\ref{figure:doorknob}d allows other body parts, arm or wrist in this example, for operation, instead of hands that might be unavailable at the moment.
Another example is a plastic bottle opener~\cite{thing_bottle_opener}, which induces the leverage. 
Different types of pen grips (e.g., \cite{thing_pen_ball}) are popular for artists as they degrade necessary wrist-power.
\textit{Actuation-reach} designs magnify parts to reach the target.
For example, light switch extension~\cite{thing_switch_extension} is useful for children, people with short stature, or situations where large furniture placed underneath makes access difficult for people using walkers.


{\textbf{(2) Constraint: Prevent operations}}
refers to designs often revert functions of \textit{actuation} designs, preventing operating objects in special contexts (e.g., cabinet lock) for people with cognitive impairments or in child-access/child-proof products.
Limiting access is another popular objective in augmentations (e.g., drawer lock~\cite{thing_drawer_lock}) favored by parents, pet owners, and caretakers of the cognitive retreat, especially for safety. 
Even for those who do not have such impairments, people label identical objects such as a series of wall switches to reduce confusion and misuse.
Common target objects contain doors, drawers, wall switches (e.g., lights and garbage disposal), or outlets that are with known risks.

{\textbf{ (3) Indication: Furnishing with visual/tactile cues}}\label{sec:accessmeta-indication}
Designs that furnish multi-modal feedback for easy identification of intention, function, or purpose by providing labels (e.g., switch labels, toggling sound); greatly benefit people with sensory impairments.
3D printed tactile graphics have gained acceptance by many people with visual impairments~\cite{buehler2015sharing}.
Built upon those principles, tactile cues provide multi-modal information to help identify functionalities in identical-looking objects, for example, 3D-printed labels in the multi-switch panel~\cite{thing_switch_identifier}.
Note that AccessMeta categories are not always mutually exclusive, as one can simultaneously furnish tactile cues and reduce motor requirements.

\subsubsection{Assistive 3D Augmentation Dictionary}
\IssueFourC{As a result of design exploration to define AccessMeta, we created an initial dictionary that contains 280 3D-printed augmentations for 52 everyday objects (e.g., handle, door, knob, book, nail clipper, knife, hair dryer, microwave, stove, table, etc.) with potential inaccessibility context, fully annotated with AccessMeta categories. 
Among 52 common object classes in AccessMeta, we found that 6 classes (i.e., handle, faucet, switch, knob, button panel, and outlet) are significant and difficult to be addressed by existing datasets with indoor scenes (e.g., ADE20K~\cite{ade20k}, COCO~\cite{COCO}) mainly due to (1) challenges caused by their small size in photos and (2) diverse types of the objects that might pose various kinds of barriers (e.g., door lever vs. knob). 
Focusing on these 6 classes (which are further divided into 21 inaccessible classes), we construct a new dataset, AccessDB/AccessReal.}
This dictionary is publicly available at \url{https://access-lens.web.app/}.


\subsection{AccessDB \& AccessReal: Dataset for Inaccessibility Detection}
\begin{figure*}[h]
\centering
\includegraphics[width=0.9\textwidth]{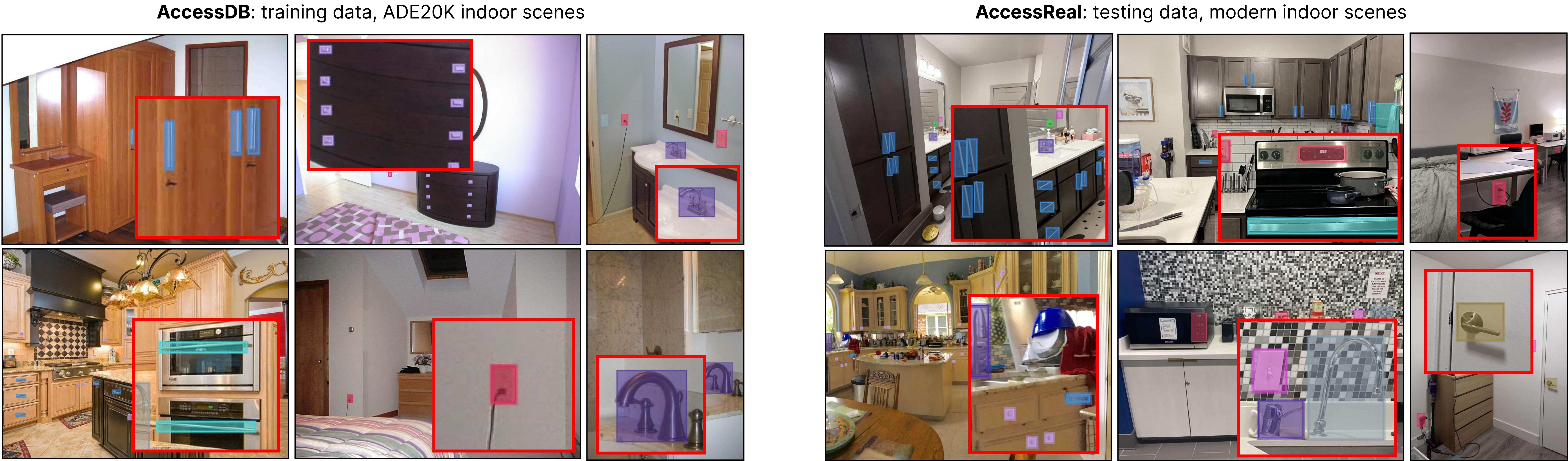}
\vspace{-3mm}
\caption{We use the AccessDB (left) and AccessReal (right) datasets to train and evaluate inaccessible-object detectors.
Images of AccessDB are sampled from the well-established ADE20K dataset~\cite{ade20k} with our re-annotation (cf. Table~\ref{tab:accessdb_accessreal_dist}).
AccessReal has high-resolution images captured by ourselves from diverse indoor scenes; we annotate these images using the same set of inaccessibility classes.
Red boxes are zoom-in regions that contain inaccessible objects.
}
\Description{The figure shows two sets of images, each in a 2-by-3 grid layout. The left set contains examples of images in the AccessDB dataset, which contains ADE20k indoor scenes, and the right set contains examples from the AccessReal dataset, which contains modern indoor scenes. All images have a box overlaying different detected inaccessible objects, and a red box inside the image that provides a zoom-in region of inaccessible objects. }
\vspace{-3mm}
\label{figure:dataset}
\end{figure*}


Auto-detecting objects with their semantics and context from camera views (e.g., \cite{rassar2022, mitsou2022enorasi, homefit_ar}) can assist visual perception for various interested groups and information processing, e.g., robotic affordance and different types of disability. 
Automation through a comprehensive dataset that provides a granularity of object classes is critical to infer necessary information from semantics.
Yet, predicting contexts from images is more complex than detecting objects and instances; \textit{object attributes} such as shapes (e.g., round, lever, cross-shaped) must relate their \textit{functional properties} (e.g., grip, twist, pinch), to be able to derive their conceptual interaction types.
Once interaction types are inferred regarding their visual and functional characteristics, those types can serve as clues to infer the original design intent as well as hidden barriers in various possible contexts.
To train and evaluate our developed inaccessibility detector, we construct two datasets: AccessDB and AccessReal.
Being built for semantic understanding of objects and their parts, 
ADE20K offers hierarchical annotations on object classes, such as \textit{closet - door - handle} and \textit{oven - door - handle}. 
AccessDB presents Inaccessibility Class (IC) to provide a nuanced understanding of diverse barriers that may manifest across various contexts, extracted from six distinct categories in ADE20K: button panels, electrical outlets, faucets, handles, knobs, and switches. The granularity of IC permits the identification of specific accessibility challenges, thus enabling tailored design solutions.
Refer to Table~\ref{tab:accessdb_accessreal_dist} in the Appendix~\ref{appendix:dataset-detail}.

\vspace{0.2cm}







\begin{figure}[h]
    \centering
\includegraphics[width=0.9\linewidth]{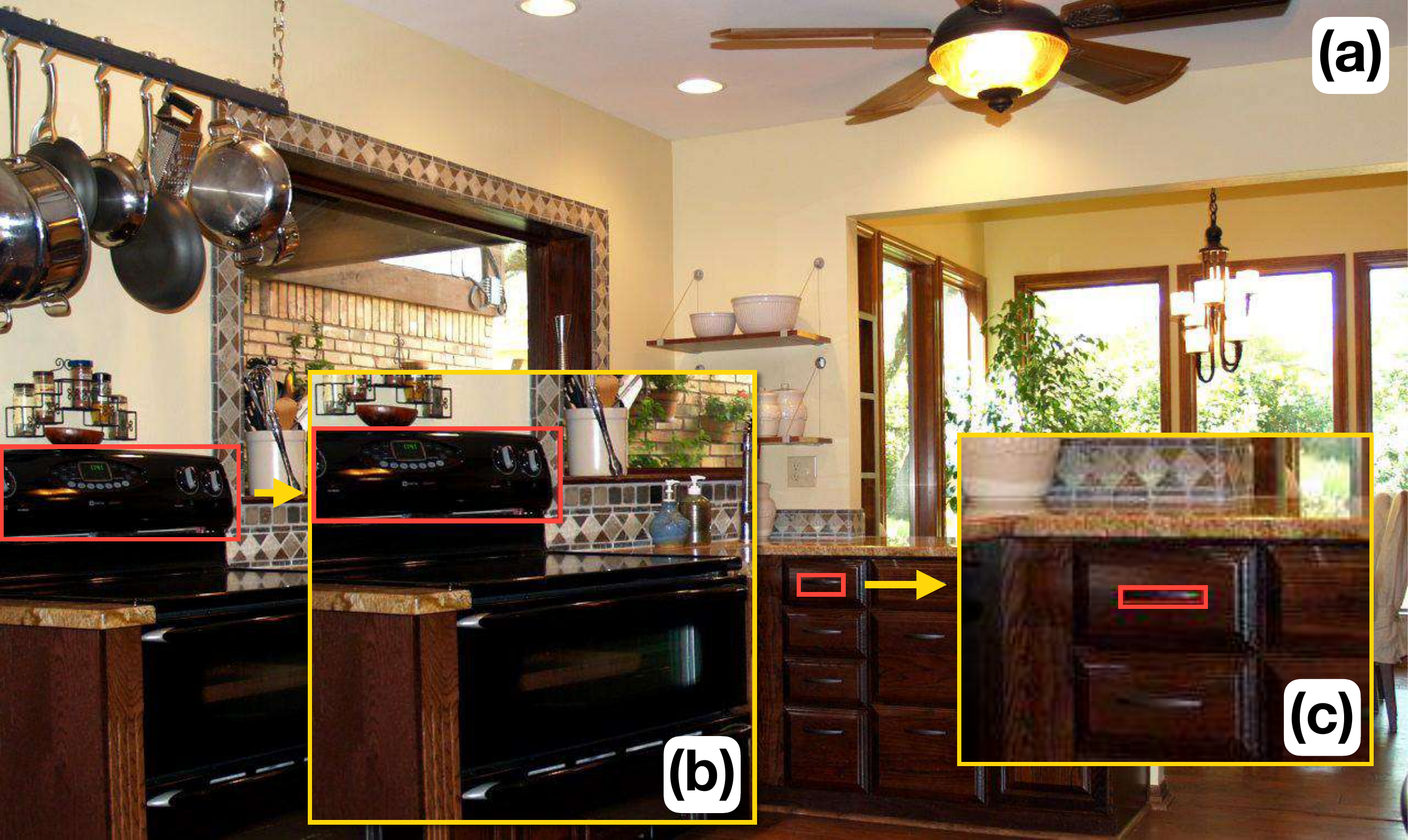}
    \caption{An example image (a) from AccessDB with two inaccessible objects: a flat button panel in a stove (b), and a handle into a drawer (c).
    These objects are often very small in the image, making annotation and detection difficult.}
    \Description{The figure shows an image of the kitchen (a), and there are two bounding boxes around the stove-top button panel and one of the drawer handles. The buttons on the panel are hard to see, even with the image the bounding box around the button panel is pointing to, image (b), which contains a zoom-in snippet of the area around the button panel. Image (c), similarly, is a zoomed-in snippet of the drawer handles, with the shape of the handles still difficult to make out.}
    \label{figure:annotation}
\end{figure}

\textbf{(1) AccessDB} is used to train inaccessibility detectors.
We derive AccessDB from ADE20K~\cite{ade20k}, which contains $>$20k images including diverse indoor scene photos with pixel-level annotations on objects and their parts.
We re-annotated objects in ADE20K for 21 predefined inaccessibility classes (ICs) in addition to ``unidentifiable'' class for extremely small sized parts.
We first select scene images sampled from ``home'', ``hotel'', ``shopping and dining rooms'', and ``workplace'', but excluded low-resolution images. 
We focus on 6 object categories that are often inaccessible (Figure~\ref{figure:accessdb_labels}): handle, faucet, switch, knob, button panel, and electric outlet. 
Three annotators are HCI experts in assistive designs, and annotators also cross-verify each other's annotations for annotation quality.
We obtained 4,976 high-resolution images exhaustively annotated with ICs as illustrated in 
Figure~\ref{figure:annotation}, which appears in extremely small regions of the image, posing a visibility challenge to detectors.

\vspace{0.1cm}
{\bf (2) AccessReal}.
Since AccessDB's images are from the ADE20K dataset which was published five years ago (as of 2023 when this research is conducted), we are motivated to curate a new dataset for evaluation by collecting photos taken in `modern' indoor scenes.
To this end, we take 42 high-resolution photos (mostly 4032$\times$3024) in diverse indoor scenes: bathroom, bedroom, kitchen, living room, and office (cf. Figure~\ref{figure:dataset}).
We annotate them w.r.t the predefined 21 ICs (see data statistics in Appendix~\ref{appendix:dataset-detail} Table~\ref{tab:accessdb_accessreal_dist}), and end up with 428 annotated objects with ICs.


\section{Evaluation}

\begin{figure*}[h]
    \centering
\includegraphics[width=\textwidth]{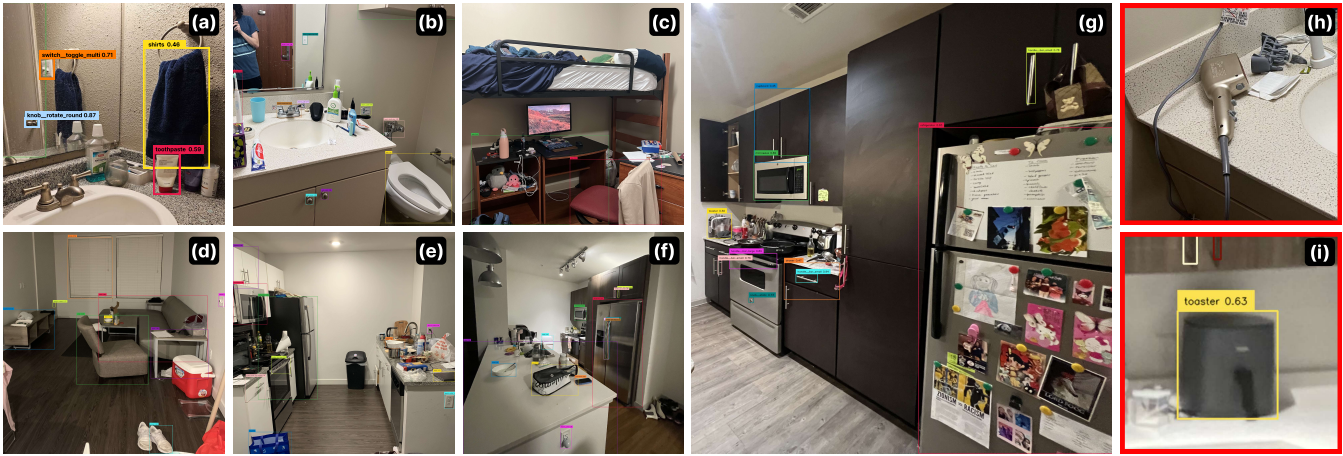}
\vspace{-5mm}
    \caption{\IssueTwoA{Examples of indoor scene photos submitted by case study participants through AccessLens. All participants took photos to show a full coverage of rooms, capturing the details as much as possible. Indoor scenes include: (a-b) bathroom, (c) bedroom, (d) living room, and (e-g) kitchen. (a-g) show bounding boxes overlaid, detected by AccessLens. Participants reported minor detection errors: undetected hair dryer (h) and air fryer misclassified as a toaster (i).}}
\Description{The figure shows examples of photos that the case study participants submitted. Image (a)
shows a bathroom with a sink, a mirror above the cabinet, and a towel rack above the sink. Image (b) shows a bathroom with a sink, an under-sink cabinet, a mirror above the cabinet, a toilet, and a towel rack above the toilet. Image (c) shows a high-loft bed with a desk underneath. There is a monitor on the desk and also drawers next to the bed. Image (d) shows a living room with a couch and table. Image (e) shows a full slot of a kitchen, with a sink counter, a strove with cabinets above, and a refrigerator. Image (f) shows a kitchen with a counter, a refrigerator, and parts of the stove with cabinets above visible. Image (g) shows another kitchen that captures a stove, cabinets around, and a refrigerator. Image (h) and image (h) show minor detection errors reported by participants. Image (h) presents an undetected hair dryer, and image (i) displays an air fryer misclassified as a toaster.}
\label{figure:case_study_input}
\end{figure*}

\subsection{\IssueTwoA{An End-to-end Pipeline}} \label{sec:case-study}


\subsubsection{Participants \& Procedure}
We conducted a holistic end-to-end study to assess: (1) capturing photos, (2) uploading photos for AI inspections, (3) viewing suggestions to address identified barriers, and (4) physically installing 3D printed results. 
We recruited six participants (U1-6) from our institution (female=4, male=2, ages 19-30) who have none to limited exposure to accessibility, except for U6 with moderate experience in technology for sign language speakers. Five (U1-5) had little or no prior experience in 3D printing, while U6 had 5+ years of experience in fabrication. None overlapped with the preliminary evaluation study participants. All studies were conducted individually. Participants first freely explored AccessLens, either on mobile or the web. They were asked to upload photos of personal space, and then select as many desired augmentations. 
Due to time constraints, we printed the chosen augmentations, except for U6 who self-printed. All attached augmentations within their environments by themselves.
Participants were asked to take photos and share the installation process, results, and thoughts. 
We concluded each study session with exit interviews. 
We took an approach similar to a contextual inquiry, with in-depth observation and interviews to gain a robust understanding of user behaviors and their motivation about specific courses of action taken, minimally intervening in the use case. All conversations and responses were transcribed and documented for analysis through coding.

\subsubsection{Results \& Finding}
Participants submitted an average of 3.7 photos/participant, totaling 22 of bathroom, bedroom, living room, and kitchen (e.g., Figure~\ref{figure:case_study_input}).

\textbf{\#1. Easy Photo-taking and Uploading.}
Although AccessLens did not provide step-by-step instructions and the facilitator minimized intervention, all naturally submitted photos of panoramic views, capturing entire rooms.
U5-6 iteratively adapted their photo-shooting strategy,\textit{``From the first try, I saw that the app detected door handles, so I ensured their visibility in subsequent photos''} (U5).
None had issues in processing photos and stated it is straightforward. 


\textbf{\#2. Learning Accessibility from Adaptation.}
Before using AccessLens, all participants expressed their lack of confidence in recognizing inaccessibility.
U5 guessed that it is possible only when obvious, e.g., seeing someone struggling in person.
U1-3 stated they \textit{``had not encountered accessibility challenges themsselves''}, and U4 found it hard \textit{``to view things from the perspective of those with accessibility issues [because I am not disabled]''.}
\\
\indent After AccessLens use, we observed elevated confidence and awareness.
\textit{``By seeing all the examples and possible solutions in my room, I now have a better understanding of potential issues and how others interact with objects differently from I do''} (U1).
U2 found the microwave button pusher~\cite{design-microwave-opener} eye-opening, since they never imagined that anyone could struggle with such simple pressing.
Most participants (U1-4, U6) testified an expansion of their perspectives; \textit{``I never thought outlets or stove buttons [could be inaccessible], since I was expecting more about people who are visually impaired or with [more serious disabilities]. I gained a new perspective that disability is such a large spectrum''} (U3).
U4 also stated, \textit{``At first I thought that the challenges would only apply to people with [diagnosed disability, but it applies to] the general population with a variety of issues, including injuries, child locks, and having busy hands.''}, confirming that users learn \textit{``potential contexts''} (U1-2, U6) through recommendations. 
U5 found being hands-free useful since the steel surfaces tend to become dirty.
AccessLens also helped U3 \& U6 redefine their experiences; \textit{``I once had a cut on my thumb, which made squeezing the toothpaste very difficult. Toothpaste squeezer seems useful (in such situations) but also on a daily basis too''} (U6).

\textbf{\#3. Perceived Accuracy of Detection.}
All participants found the automated detection accurate, expressing confidence in interpreting the results.
U3 was concerned about messy rooms but was impressed by the detector performance that captured objects successfully even from cluttered scenes.
U6 found that even a small reflection of the door knob in a mirror was correctly detected.
AccessLens was thought accurate only except for U1's hair dryer, possibly due to its uncommon design (Figure~\ref{figure:case_study_input}h), and
U4's air fryer is seen as a toaster (Figure~\ref{figure:case_study_input}i).
All were thought minor and did not affect participants' trust in overall detection results.

\textbf{\#4. AccessMeta and Dictionary Supporting Exploration.}
Participants appreciated AccessLens' presentations, organized by the detected objects and related issues with AccessMeta.
Participants (U2-3, U5) found the dictionary explorer, which shows all possible designs useful.
\textit{``Before reading the dictionary, I was not aware of child safety and how they related to accessibility, but the dictionary helped me learn potentially dangerous aspects of objects and how to mitigate them''} (U1).
U3 perceived the variety of the dictionary as very useful for browsing especially \textit{``when moving to a new place, remodeling, or choosing new appliances''}.
U6 imagined augmenting standard spaces with various needs; \textit{``The standard apartment's equipment is not designed for specific needs. People will find it very useful to augment their everyday environment with specific needs in mind''}.

\begin{figure}[h]
\centering
\includegraphics[width=0.48\textwidth]{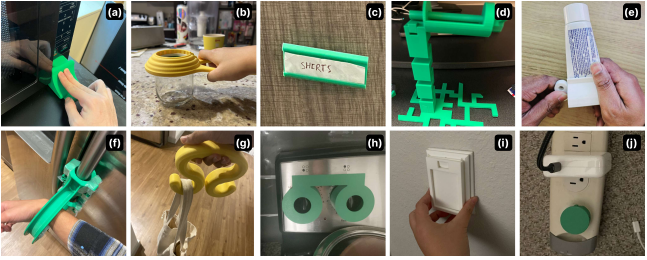}
\caption{\IssueTwoA{Retrofitting 3D-printed augmentations by study participants: (a) microwave opener, (b) jar opener, (c) drawer label holder, (d-e) toothpaste squeezer, (f) hands-free door opener, (g) bag holder, (h) stove knob cover, (i-j) outlet cover.}}
\Description{This figure presents 10 images illustrating augmentations by case study participants in various use cases: (a)using a microwave opener to press a microwave's button, (b) opening a jar with a jar opener, (c) a drawer equipped with a label holder, (d) a toothpaste squeezer, (e) squeezing toothpaste with a toothpaste squeezer, (f) opening refrigerator doors using a hands-free door opener, (g) lifting a heavy bag with a bag holder, (h) a stove knob covered with a knob cover, (i) using a sliding outlet cover to over the outlet, (j) an outlet covered with a circular outlet cover outlet.}
\vspace{-5mm}
\label{figure:retrofit}
\end{figure}

\textbf{\#5. Different Motivations to Adopt AccessMeta Recommendations.}
Each participant selected 2-4 augmentations, such as a hands-free opener for large door handles, electric outlet covers, jar openers, stove knob protectors, etc (example retrofitting results seen in Figure~\ref{figure:retrofit}).
Their selection criteria varied: frequency of use (U1, U6), assistance when alone (U2), safety (U3, U5), practicality, and sheer interest (U4).
Some could still find useful designs through an inductive process, not necessarily having the images; 
\textit{``I know my parents or grandparents struggle using, such as a toenail clipper as they don’t have enough back flexibility. It’s nice to have the option to look at suggestions [without having the images] of their houses''} (U2).
We imagine AccessLens' advanced feature for expanded recommendations. If the contextual disabilities are known through the user's previous choices of recommended adaptations, AccessLens can fetch common objects that present similar barriers.

\textbf{\#6. Low-cost Upgrades through Retrofitting but Need to Handle Uncertainty.}
Participants found 3D-printed upgrades easy and cost-effective.
All were able to install the augmentations without any help and did not face major difficulties, spending a maximum of a few minutes when designs required assembly. 
Many designs on Thingiverse are versatile and modular, often in standard dimensions or using screws for a tight fit.
Participants found standalone designs (e.g., bag holders, knob covers) were easy to utilize. For example, U3 found that the stove knob lock fit perfectly, and found it useful for safety when children or cats are around.
With assembly, participants were actively involved in the adaptation.
U1 found that the microwave door opener~\cite{design-microwave-opener} is slightly taller, so they tilted the microwave up to match the height. U4 and U5 did not have screws to put parts of the hands-free door opener~\cite{design-hands-free-opener}, but still made it work by installing it using tape. 
For designs that need assembly, three participants (U1-3) thought having a step-by-step guide would be beneficial.
While all successfully adopted designs, some reported dimensional challenges; U3's outlet covers did not fit so they had to put it over without fixation. U5's hands-free fridge opener was loose and slid, failing to stay at arm height.
We consider integrating well-established customization tools focusing on a fit, e.g., ~\cite{kim2017understanding, hofmann2018greater} and auto-measurement~\cite{customizAR}.




\textbf{\#6. Additional Suggestions.}
Overall, participants were satisfied and willing to continue using AccessLens.
Three participants (U3-5) suggested a detailed description for augmentations clarifying the functionality and objectives on the app without redirecting to the design page.
U1-2 and U4 also mentioned that showing the required materials (e.g., screws, tape, clips) would be helpful for users to make choices based on complexity and material availability.
U6 also hoped to see an animated preview of how the augmentation could change the interaction.

\subsection{Expert Feedback about User Experience}

\subsubsection{Participants} 
The expert feedback session was conducted to understand how AccessLens can support users to raise awareness about accessibility. We engaged two professionals (E1-2) with 10+ years of expertise in accessibility research and teaching access computing.
E1's expertise lies in robotics for people with movement disabilities and/or chronic conditions (e.g., people with Parkinson's disease, and freezing of gait), and E2's expertise is in assistive visual perception for the visually impaired through systems for human-AI interaction.
We sought their qualitative opinions about various topics of interest: user engagement, system functionality, empowerment in decision-making, alignment with standards, usability, potential impact, and future developments. 

\subsubsection{Findings}
Both acknowledged the tool's diverse and relevant suggestions, particularly for \textit{``raising awareness of accessibility issues, aiding those without specialized accessibility knowledge''} (E1).
Yet, E1 expressed concerns about non-experts due to the absence of a clear description of relevant accessibility issues for diagnosed disabilities. 
While the system provides real examples and suggestions for environmental modification facilitating users' perception of various possible contexts indirectly, it lacks \textit{``explicit explanations''}, potentially hindering informed decision-making. 
E2 commented about possible design conflicts, \textit{``if multiple people residing in the space with different accessibility needs, solutions could be in conflict with each other, or the design needs to be combined to satisfy multiple needs''}. 
AccessLens needs more targeted customization and alignment with \textit{public accessibility standards}, E1 added.
Similarly, while appreciating the system's ability to identify numerous relevant objects, E2 suggested incorporating design parameters, including configuration/layout of the environment (e.g., the width of a hallway) and interaction/spacing between objects (e.g., the distance between switch and floor), which we find incorporating physical assertion of adaptive designs~\cite{hofmann2018greater} critical.
E1 sees long-term benefits, particularly for growing 3D printing communities but with limited accessibility knowledge. 
E2 also proposed allowing users to input disability types to prioritize suggestions and emphasize the importance of customizing solutions in mind. 
E2 imagined crowdsourcing for more examples and an onboarding feature for new users to enhance utility.
In summary, both experts recognize AccessLens's potential to engage inexperienced users.
Encompassing customization support to accommodate various physical dynamics, guidance, and user-defined disability prioritization at the input stage can further improve AccessLens.

\begin{figure*}[h]
\centering
\vspace{-0.5cm}
\includegraphics[width=0.8\textwidth]{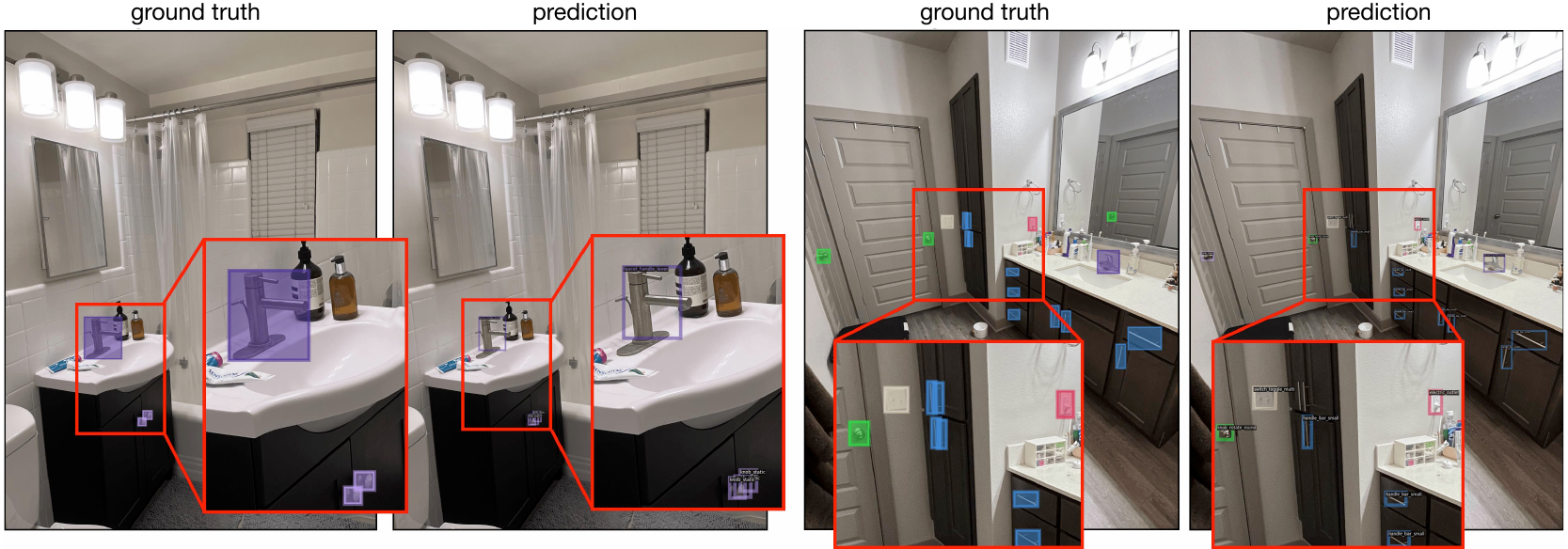}
\caption{Ground-truth and detection results of our inaccessible-object detector on two example images in AccessReal. For brevity, we omit IC labels (and detection confidence scores) in ground truth but present only labels for detection boxes. A visual examination of the results reveals that our detector exhibits a decent capability for identifying inaccessible objects.
}
\Description{The figure displays two sets of images, each set featuring different bathrooms. In each pair, one image shows the `ground truth' - the actual layout of the bathroom without labels. The other image in the pair shows `detection results' from our inaccessible object detector, highlighting objects with labeled boxes.}
\vspace{-5mm}
\label{figure:visual-results}
\end{figure*}

\vspace{-1mm}
\subsection{\IssueTwoB{AccessMeta's Acceptance}}

\subsubsection{Procedure} 
To assess the acceptance of AccessMeta, we conducted an independent study on human annotators' perception and consensus on AccessMeta and a fully annotated dictionary of 280 3D augmentations by the research team.
Using Amazon Mechanical Turk, we designed tasks to assess how well the general public understands AccessMeta's classification criteria.
In each HIT, annotators engage with one 3D augmentation and categorize it under one of the three high-level categories from AccessMeta: `actuation', `constraint', and `indication'. These categories are further organized into five sub-categories: `actuation-reach', `actuation-operation', `constraint', `indication-visual', and `indication-tactile'. 
An additional `others' option was provided for a custom label, if any.
Annotators opened a provided URL (e.g., Thingiverse) and chose the label(s) that best describe the augmentation. To avoid potential bias, we did not provide any image references. Instead, annotators are provided textual descriptions from AccessMeta. 
We consider a HIT submission acceptable in acceptable scenarios: an annotator (1) correctly identifies the specific label (e.g., `actuation-reach'), (2) chooses a subcategory under the correct high-level category (e.g., selecting `actuation-operation' for an `actuation' design), 
(3) chooses multiple labels within the correct high-level category, and (4) selects `others' with a reasonable custom label.

Submitted HITs were first reviewed by the second author and were subject to rejection only when they fell under four cases: 
(1) \textit{all} annotations provided by a single annotator for different design entries were identical and incorrect (all HITs would be rejected);
(2) an annotator selected `others' but provided irrelevant tags such as too generic comments (`good design'), unrelated phrases (`We and our 814 partners'), or simply copied the full title or description of the design; 
(3) all responses submitted by a single annotator were incorrect and completed in less than 40 seconds (threshold decided from the test run), which indicates insufficient time to complete the task;
(4) a single annotator submitted more than 100 HITs, any responses beyond the 100-HIT limit would be rejected to ensure diversity in results. 
Results were again verified by the first author. 
\textit{N}=515 HITs were rejected and republished for re-annotation.
A worker was paid \$0.05 per HIT, and one worker submitted 16.8 annotations on average. 

\vspace{-3mm}
\subsubsection{Results \& Findings}
Three different annotations were collected for each of the 280 designs, eventually obtaining 839 valid annotations from 83 workers.
The median completion time was 6.8 minutes (8 sec. to 30 min., std = 6.8 min.)

\textbf{Acceptability.} 
If workers' annotations matched the ground truths of three main classes, they were marked as success, otherwise, failure.
Accuracy was analyzed by the ratio of correct annotations over total annotations obtained (N=839) across 280 designs.
Annotators showed 83\% match (\textit{N}=697), implying fair acceptance of AccessMeta. 
For about 20\% of correct annotations, workers' selection of subcategories varied within a category, e.g., `actuation-reach' instead of `actuation-operation', possibly due to the versatile nature of assistive designs. 
As discussed earlier,
AccessMeta subcategories are not always mutually exclusive. For instance, tactile indications often provide visual cues, and extensions to help reach items can also facilitate alternative or smoother operation.

\textbf{Category Expansion by Annotator-Adaptation.}
About 98\% of annotations were made from AccessMeta categories.
Despite not many (1.8\%), 10 workers selected the 'others' option for 13 designs.
Three new classes emerged, mostly for designs labeled as `actuation-operation' (e.g., hands-free book holder~\cite{design-book-holder}, ziploc back holder~\cite{design-ziploc-holder}, cup holder attachable to the sofa~\cite{design-sofa-cup-holder}): `holder' (\textit{N}=6), `stabilizer' (\textit{N}=2), and `support' (\textit{N}=2).  
Annotators also suggested `protector' (\textit{N}=2) and `safety' (\textit{N}=1) for child-proof designs---a child finger protector for drawers~\cite{design-drawer-protector} and a sharp corner protector~\cite{design-table-protector}, respectively, which are currently defined as `constraint'.
Growing in complexity with diverse contexts and objects, we perceive AccessMeta to serve as a platform to expand through the collective input for more diverse \& inclusive classifications. Future work could involve mechanisms for reports/suggestions from stakeholders and designers for adaptive solutions.

\vspace{-1.5mm}
\subsection{\IssueTwoC{AccessDB Detector Performance}}
 
Our approach allows adapting any state-of-the-art detector architectures (e.g., GroundingDINO~\cite{groundingdino} and RetinaNet~\cite{retinanet}; details in Appendix~\ref{appendix}). 
Figure~\ref{figure:visual-results} displays example detection results on AccessReal images, showing good qualitative performance in detecting small inaccessible objects. 
In evaluation, (Section~\ref{sec:case-study}), all participants showed solid trust in our detector's performance.
Detection result visualizations for sample images in AccessReal (Figure~\ref{figure:visual-results}) also show that the detector accurately captures small objects.
\IssueFourB{AccessDB and AccessReal datasets are open-sourced at \url{https://access-lens.web.app/} to foster future research. While our work used one of the state-of-arts, any modules can be trained on our dataset. For technical specifications, refer to Appendix~\ref{appendix}.}

\color{black}

\vspace{-1.5mm}
\section{Discussion \& Future work}
\subsection{\IssueFourA{Collective Disability Accommodations}}

Engaging with a building ADA coordinator at our institution sheds light on a collective effort in identifying/reporting.
The ADA coordinator admitted that many staff lack accessibility expertise, so they hire external accessibility specialists to address issues on demand.
Encouraging citizen science within our initiative could mirror successful collective intelligence models like Project Sidewalk~\cite{project-sidewalk}. 
By adopting a reporting system where individuals contribute to accessibility assessment within commons, accommodating potentially inaccessible physical environments but have \textit{not yet} discovered by people with diagnosed disabilities before they encounter barriers. 
Experts' recommendations to input disability types and validate possible conflicts can be applied to seek AccessLens at scale.
\vspace{-1.5mm}
\subsection{Is AccessLens A Disability Dongle?}

The term `disability dongles' has emerged to criticize endeavors that employ innovative technologies but fail to address genuine accessibility needs~\cite{disability-dongle}, often targeting industry products that exploit accessibility concepts for superficial gains.  
AccessLens builds upon prior work on understanding and addressing existing issues of recognizing accessibility barriers~\cite{rassar2022}.
Our approach utilizes the state-of-the-art technologies that are becoming more and more available to innovate an individual's life (e.g.,~\cite{baudisch2017personal}), thereby assisting users make informed decision-making. 
Moving one step forward, AccessLens broadens its impact to a wider audience who do not experience diagnosed disability, provoking discourse about disabled contexts~\cite{microsoft_inclusive_design}. 
3D printed and/or DIY solutions already become the major efforts made by numerous disabled selves, stakeholders, and altruistic enthusiasts~\cite{chen2016reprise, buehler2015sharing}. As AccessLens promises to incorporate more options for store-bought solutions and the industrial design industry (e.g., \cite{homefit_ar}), we anticipate more collaborative efforts across disciplines to advance people's quality of life using technology as we detail in the following section.

\color{black}
\vspace{-1.5mm}
\subsection{\IssueTwoB{Expanding to 3rd-Party Solutions}}

AccessMeta links the object types with their needed interaction, seeking solutions that might alter interaction types (e.g., grab-rotate-to-open vs. push-open).
Once detected, we see the future of AccessMeta and the dictionary expanding the search for similarly-functioning 3rd-party alternatives, such as buying door lever replacements from hardware stores or online markets.
While some simple replacements like doorknobs might be as cheap as 3D printing, more complex fixtures such as refrigerator handles (as in Figure~\ref{figure:retrofit}) are not trivial, necessitating the disassembly or replacement of the whole appliances.
Although our study participants agreed on the less mental burdens with AccessLens recommendations, some were inclined towards store-bought products as they have gone through market testing (U3), given their perceived affordability and time cost for customization (U5).
As AccessLens provides direct recommendations compared to \textit{``for store-bought ones, I might have to look for products on my own that solve the highlighted challenge for detected objects"} (U2), offering users more options upon various rationale, control for materials (U5), easy-fix and remix (U6).

\vspace{-1.5mm}
\subsection{3D Model Customization}
3D printing is a promising solution for custom adaptive interfaces to meet unique needs. 
One notable example is auto-filling numerics into parametric 3D designs~\cite{customizAR} and in creating various branches of augmentations upon user's needs to adapt common household items~\cite{chen2016reprise}. 
The current AccessLens prioritizes inaccessibility detection and assistive augmentation recommendations. 
As our work has been focusing on increasing awareness and low-cost solutions, dealing with fit~\cite{kim2017understanding} and parametric customization was considered orthogonal.
However, we recognize the potential synergy with existing works facilitating customization (e.g.,~\cite{hofmann2018greater}), starting from the auto-detection and selection of a suitable design and culminating in real-world applications. 
We can further empower individuals to take proactive steps toward creating inclusive environments. 
\vspace{-2mm}
\subsection{Expanding AccessDB \& AccessReal Dataset, Populating AccessMeta}
This work provides two challenging datasets, AcessDB and AccessReal for inaccessibility detectors.
Communities' interest in inclusive designs has grown, and advances to automate everyday surroundings (e.g., smart switches, thermostats with touch screens) create new challenges; touch screens often lack tactile feedback for people with visual impairments and can present more challenges for the elderly).
To scale the dataset, this work elaborated on the re-annotation strategy of AccessDB in detail at our dataset website.
We believe that the AccessMeta pipeline should remain open-ended and adaptable to accommodate emerging needs and novel designs. 
One approach to expanding the AccessMeta pipeline is involving a community in reporting problems and suggesting additional metadata categories. We can ensure that the system remains responsive to real-world needs, identifying new challenges.

\vspace{-1.5mm}
\subsection{Can AccessLens Promote Altruism?}
\IssueFourA{We envision the use of AccessLens will help people become more aware of implicit inaccessibility and more actively engaged in improving access in public spaces, such as lecture rooms and shared dormitory community rooms.
We have not observed positive behavioral changes in participants beyond the lab.}
We plan to conduct a deployment study to evaluate whether AccessLens raises people's awareness and encourages collective actions, similar to how altruism motivates voluntary sharing of designers online for free. 
Expert interviews from diverse domains, including HCI, accessibility, visualization, and citizen science, will be conducted to critique the user interface and study design systematically, ensuring unbiased evaluation of AccessLens compared to other existing tools.
\vspace{-2mm}
\section{conclusion}

AccessLens provides an end-user tool that helps users without diagnosed disabilities or prior experiences in accessibility assess the accessibility challenges.
We adopted object detection techniques to train inaccessible-object detectors on our novel dataset AccessDB. 
On our collected dataset AccessReal which consists of images of modern indoor scenes, we show that our detector can detect inaccessible-objects well.
We designed AccessMeta to link inaccessibility classes to keywords of 3D assistive augmentations.
Through two rounds of holistic evaluation with inexperienced users, we demonstrate the effectiveness of AccessLens in raising awareness and proactiveness in improving indoor accessibility.

\vspace{-2mm}
\begin{acks}
We extend our appreciation to Dr. Momona Yamagami and Dr. Anhong Guo for their invaluable feedback and discussions on AccessLens' potential and improvements. We are grateful to Dr. Megan Hofmann for her expert consultation on access computing \& disability dongles. Shu Kong is supported by University of Macau (SRG2023-00044-FST) \& Institute of Collaborative Innovation.
\end{acks}

\bibliographystyle{ACM-Reference-Format}
\bibliography{main}

\appendix


\section{Dataset Details} \label{appendix:dataset-detail}

AccessDB/Real comprises around 10k re-annotated objects across 21 ICs. Further details regarding the breakdown of our dataset can be found in Table~\ref{tab:accessdb_accessreal_dist}.

\begin{table}[!h]
\small
\centering
\begin{tabular}{llll}
\toprule
\multicolumn{1}{c}{\textbf{id}} & \multicolumn{1}{c}{\textbf{inaccessibility class}} & \multicolumn{1}{c}{\textbf{AccessDB}} & \multicolumn{1}{c}{\textbf{AccessReal}} \\
\midrule
1                               & button\_panel\_push\_buttons      & 83                                    & 14                                      \\
2                               & button\_panel\_turn\_handle       & 165                                   & 8                                       \\
3                               & electric\_outlet                  & 1,382                                  & 33                                      \\
4                               & faucet\_faucet\_only              & 169                                   & 3                                       \\
5                               & faucet\_handle\_lever             & 351                                   & 13                                      \\
6                               & faucet\_pull\_tiny\_knob          & 29                                    & 0                                       \\
7                               & faucet\_rotate\_cross             & 86                                    & 0                                       \\
8                               & faucet\_rotate\_knob              & 96                                    & 0                                       \\
9                               & handle\_bar\_large                & 375                                   & 19                                      \\
10                              & handle\_bar\_small                & 1,712                                  & 191                                     \\
11                              & handle\_cup\_handle               & 243                                   & 31                                      \\
12                              & handle\_drop\_pull                & 491                                   & 0                                       \\
13                              & handle\_flush\_pull               & 43                                    & 0                                       \\
14                              & handle\_lever                     & 211                                   & 10                                      \\
15                              & handle\_pull                      & 289                                   & 14                                      \\
16                              & knob\_rotate\_round               & 205                                   & 26                                      \\
17                              & knob\_static                      & 3,026                                  & 38                                      \\
18                              & switch\_rocker\_multi             & 84                                    & 3                                       \\
19                              & switch\_rocker\_single            & 57                                    & 4                                       \\
20                              & switch\_toggle\_multi             & 103                                   & 8                                       \\
21                              & switch\_toggle\_single            & 115                                   & 13     
              \\

22                              & unidentifiable            & 724                                   & 0     
              \\
              \midrule
                                & \textbf{total}                             & \textbf{10,039}                                 & \textbf{428}     \\
\bottomrule
\end{tabular}
\caption{Counts of annotated objects per inaccessibility classes in AccessDB and AccessReal datasets. 
There are 21 inaccessibility classes plus an ``unidentifiable''.
AccessDB and AccessReal contain 2,388 and 42 indoor scene images, respectively. We use AccessDB for training and validation, and AccessReal as the testing set for evaluation.}
\Description{The table shows the number of objects for AccessDB and AccessReal in the inaccessibility class of each row. }
\label{tab:accessdb_accessreal_dist}
\vspace{-5mm}
\end{table}

\section{Example assistive augmentations}
\begin{figure*}[!h]
    \centering
    \vspace{-2mm}
    \includegraphics[width=\linewidth]{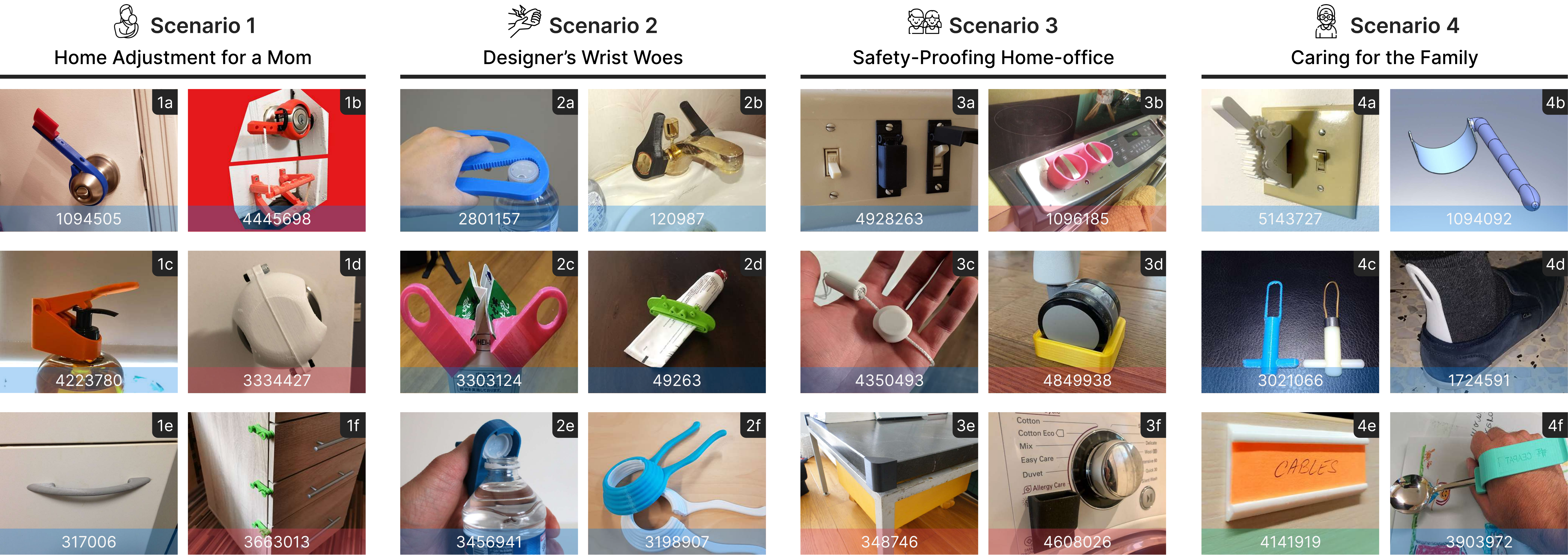}
    \caption{Augmentations recommended by AccessLens for four example scenarios: (1) home adjustment for a new mom, (2) designers who has chronic wrist pain, (3) safe-proofing home office, and (4) caring for the family member who is an older adult. Each design has a \textit{thing\_id} at the bottom label, and the design page locates at https://www.thingiverse.com/thing:\textit{thing\_id}. Labels indicate blue: actuation, red: constraint, green: indication.}
    \vspace{-2mm}
    \Description{The figure shows six photos of 3D augmentation suggestions for four different scenarios. The first scenario, `Home Adjustment for a Mom', has suggestions that include a lever door extension, a hands-free foot opener, a bottle pump dispenser, a knob cover preventing child access, a handle to stick onto cabinets or drawers, and child-safety drawer locks. Scenario 2, `Designer's Wrist Woes', is paired with suggestions including a bottle opener, a knob extension for a faucet, a milk carton opener, a toothpaste squeezer, and a jar opener. The third scenario, `Safe-proofing Home-office', has light switch covers, stove knob protectors, a window blinds cord hanger, a rolling chair wheel stopper, a washing machine button cover, and a table corner protector. Scenario 4, `Caring for the Family', is paired up with suggestions including a gear light switch extension, a tool to put on socks, a shoe horn, a button hook to aid buttoning with one hand, a cabinet label, and a clamp to hold eating utensils.}
    \label{figure:scenarios}
\vspace{-0mm}
\end{figure*}

We introduce example augmentations with possible user walkthrough of AccessLens in various user contexts in Figure~\ref{figure:scenarios}.

\textbf{Scenario \#1: Home Adjustment for a Mom.}
Mark's wife struggles to care for a 6-month-old baby and do housework. Mark wanted to upgrade his home. He scanned the rooms using AccessLens to get recommendations for common objects such as doorknobs (Figure~\ref{figure:scenarios} 1a), water faucets, and lower drawers (1e-f).
It also proposed an arm-activated handle (1a) and a foot-open door handle (1b). It suggested an one-handed hand soap dispenser which he did not notice was hard to use but useful if it applied to the baby lotion bottle for one-hand dispense (1c).
Mark now understands what could be inaccessible with arms occupied by a baby.

\textbf{Scenario \#2: Designer's Wrist Woes.}
Enter Kathy, a designer grappling with chronic wrist pain that hindered daily tasks. AccessLens scrutinized her kitchen, finding jar lids (2f), plastic bottles (2a, c), and milk cartons (2c) were known to be inaccessible. Provided with custom-designed openers tailored to each item, she feels relief for her weakened wrists. Kathy also opted for 3D-printed extensions for her faucets (2b) so they open by push not by grab-rotating. Receiving a snapshot of the bathroom, AccessLens recommended a toothpaste squeezer (2d) that alleviates the strain. 

\textbf{Scenario \#3: Safety-Proofing Home-office.}
Arjun, a single individual is preparing to host a home party at his home studio with a safe environment for families with young children. He reviewed the home using AccessLens, identifying potential risks that we never were aware of. Recommendations ranged from switch covers (3a) to prevent sink grinder accidents, child safety stove knobs (3b), machine button covers (3f), and a cord hanger (3c) to prevent the blind cords hazard. He also prepared several 3D-printed bumpers (3e) to be attached to sharp edges, and drawer locks (1f), which can be also useful for Julie. Seemingly innocuous office chair wheels (3d) were also covered by AccessLens.

\textbf{Scenario \#4: Caring for the Family.}
Mia is a devoted daughter and caretaker of her elderly mother, who is increasingly lacking mobility capacity. 
AccessLens suggested specialized tools designed to facilitate daily routines: a sock aid (4b) to help avoid bending, a button hook (4c) to simplify fastening shirts, and an extended shoe horn (4d). Additionally, AccessLens recommended a switch extension (4a), allowing her mother to operate it easily without precise hand movements or while on home medical equipment. Mia also used drawer labels (4e) for her bath products for easy identification using larger texts.
Enabling greater comfort and independence in managing daily tasks, Mia also found those are overall accessible for her young child.

\color{black}

\section{Detector Performance} \label{appendix}

\begin{table}[!h]
\begin{tabular}{llll}
\toprule
\multicolumn{1}{c}{\textbf{id}} & \multicolumn{1}{c}{\textbf{inaccessibility class}} & \multicolumn{1}{c}{\textbf{AccessDB}} & \multicolumn{1}{c}{\textbf{AccessReal}} \\
\toprule
1  & button\_panel\_push\_buttons & 13.91        & 11.43          \\
2  & button\_panel\_turn\_handle  & 26.48        & 7.72           \\
3  & electric\_outlet             & 29.94        & 16.65          \\
4  & faucet\_faucet\_only         & 21.36        & 4.90           \\
5  & faucet\_handle\_lever        & 29.92        & 12.85          \\
6  & faucet\_pull\_tiny\_knob     & 38.52        & n/a            \\
7  & faucet\_rotate\_cross        & 34.84        & n/a            \\
8  & faucet\_rotate\_knob         & 32.92        & n/a            \\
9  & handle\_bar\_large           & 13.04        & 5.7            \\
10 & handle\_bar\_small           & 16.78        & 10.37          \\
11 & handle\_cup\_handle          & 3.21         & 0.04           \\
12 & handle\_drop\_pull           & 27.99        & n/a            \\
13 & handle\_flush\_pull          & 0.80         & n/a            \\
14 & handle\_lever                & 12.38        & 15.71          \\
15 & handle\_pull                 & 9.40         & 1.03           \\
16 & knob\_rotate\_round          & 29.08        & 23.01          \\
17 & knob\_static                 & 16.34        & 2.26           \\
18 & switch\_rocker\_multi        & 14.31        & 23.50          \\
19 & switch\_rocker\_single       & 1.53         & 2.02           \\
20 & switch\_toggle\_multi        & 31.36        & 64.21          \\
21 & switch\_toggle\_single       & 10.52       & 36.20         \\
\midrule
& average & 18.85 & 14.86 \\
\bottomrule

\end{tabular}
\caption{Breakdown results of our inaccessible-object detector on  AccessDB validation set and AccessReal. Performance is measured by AP for each inaccessibility class.
AP metrics on AccessDB are generally higher than AccessReal, showing a reasonable domain gap. Yet, on some inaccessibility classes such as switch\_toggle\_single and switch\_toggle\_multi, AP metrics on AccessReal are higher, presumably because images of AccessReal are higher in resolution that these small inaccessible objects are clearer and easier to detect than AccessDB images.
}
\Description{This table presents the detector performance, measured by AP, on the AccessDB and AccessReal datasets for each of the 21 inaccessibility classes that are presented in individual rows. }
\vspace{5mm}
\label{tab:performance_class}
\end{table}

{
\setlength{\tabcolsep}{1.9em} 
\begin{table}[!h]
\centering
\begin{tabular}{lcccccccccccccccccccc}
\toprule
&  {\bf mAP} & {\bf AP$_{50}$} & {\bf AP$_{75}$} 
\\
\midrule 
AccessDB
& 18.85
& 33.41
& 19.03 
\\
AccessReal
& 14.86
& 28.24
& 11.55
\\
\bottomrule
\end{tabular}
\caption{We evaluate our inaccessible-object detector (based on the RetinaNet architecture~\cite{retinanet}) on the validation set of AccessDB, and the AccessReal (as the testing set). 
Quantitative results show a clear domain gap between the two datasets;
visual results in Figure~\ref{figure:visual-results} demonstrate that our detector (trained on AccessDB's training set) can detect inaccessible objects quite well in AccessReal, representing modern indoor scenes.
}
\Description{This table's header row is mAP, AP50, and AP75. As a header column, it has AccessDB and AccessReal.}
\label{tab:baseline} 
\end{table}
}


\subsection{Evaluation Metrics.}
The literature of object detection commonly uses the standard metric of mean Average Precision (mAP) at interaction-over-union (IoU) thresholds ranging from 0.5 to 0.95, with a step size 0.05~\cite{Lin2014MicrosoftCC}.
We use mAP as the primary metric.
Following other prior works~\cite{dwibedi2017cut, ammirato2018target, mercier2021deep}, we also report performance with respect to the metrics of AP$_{50}$ and AP$_{75}$~\cite{hodavn2019photorealistic}, meaning the Average Precision (AP) at IoU threshold 0.5 and 0.75, respectively.

\subsection{Training a detector with AccessDB}
AccessLens supports detection for all object classes in the 3D assistive augmentation dictionary and ICs.
Although any detector structure can be chosen, we utilized two different state-of-the-art methods, RetinaNet~\cite{retinanet} for ICs with training on AccessDB, and GroundingDINO~\cite{groundingdino} for zero-shot detection without training for more common classes (e.g., sofa, table, cup, etc.)
Specifically, we trained RetinaNet~\cite{retinanet} with ResNet-50-FPN backbone with 3x LR schedule, implemented by detectron2~\cite{wu2019detectron2}.
In training, we employed COCO pretrained weights retrieved from Model Zoo of detectron2.
As Figure~\ref{figure:accessdb_labels} illustrates, AccessDB contains 21 inaccessibility classes, and one extra class, unidentifiable instances due to their extremely small size to identify with human eyes.
For training and validation of the detector, we randomly split the dataset, 85\% training and 15\% validation (2,029 and 359 images for training and validation, respectively). 
We used the AccessReal dataset for testing (42 images) to understand and compare how the detector works on AccessDB and more high-resolution images in AccessReal.
For the `unidentifiable' class, we still included it as an individual class in training but did not use it for evaluation. 
This is because, `unidentifiable' objects are still in the 6 categories of our interests, so those might have overlapping visual features with other inaccessibility classes that the human eye could not capture due to the blurry images. 
By treating it as one class in training a detector, we can avoid unwanted penalizing of the other classes' correct predictions.


\subsection{Detector Analysis}
Evaluation of the detector on validation and test sets was performed per each epoch.
The detector achieved its best performance for the AccessReal dataset after around 51 epochs, yielding an mAP of 18.85 for the validation set and 14.86 for the test set. Additional performance metrics are provided in Table~\ref{tab:baseline}.
AccessDB validation set showed the best mAP (19.86) at epoch 61, but after 51 epochs the detector started overfitting to AccessDB, resulting in the lower mAP (13.36) for AccessReal.
Even though AccessDB and AccessReal both contain real-world indoor images, we could still see the domain gap between the two as the detector shows about 4 less mAP. 
We attribute this performance difference, in part, to the significantly higher resolution of images in AccessReal, which poses a challenge for a detector primarily trained on smaller images. Furthermore, AccessDB inherently exhibits a long-tailed distribution in terms of class counts
(Detailed breakdown of the number of classes is described in Table~\ref{tab:accessdb_accessreal_dist}). This distribution presents an additional challenge to the detector, particularly when recognizing classes with a relatively small number of objects, which may not provide sufficient data for the model to learn distinctive visual features.
Despite the challenges, visual results created by our detector (Figure~\ref{figure:visual-results}) showcase its ability to perform well on high-resolution indoor images. 
In the zoomed regions of Figure~\ref{figure:visual-results} (second and fourth images), results show that the detector successfully recognized our interested objects, including knob\_rotate\_round, faucet\_handle\_lever, electric\_outlet, and handle\_bar\_small. 
Table~\ref{tab:performance_class} provides a breakdown of mAP for each inaccessibility class.
The average mAP indicates that, as a whole, the detector performs better on AccessDB compared to AccessReal.
However, it's worth noting that the detector exhibits superior performance on AccessReal for certain classes, such as switch\_toggle\_multi, switch\_toggle\_single, and handle\_lever. 
We hypothesize that for these classes, AccessReal may offer clearer object representations or exhibit fewer visual variations, possibly due to its smaller sample size, thereby contributing to improved detection accuracy.

\end{document}